\documentclass[preprint,12pt]{elsarticle}

\usepackage{amssymb}
\usepackage{amsmath}
\usepackage{subfig}
\usepackage{multirow}
\usepackage{makecell}
\usepackage{algorithm}
\usepackage{algorithmic}
\usepackage{graphicx}
\usepackage{xcolor}
\usepackage{float}
\usepackage{microtype}
\usepackage{rotating}
\usepackage{placeins}
\usepackage{siunitx}
\usepackage{url}
\journal{Neurocomputing}

\begin{document}

\begin{frontmatter}

\title{CAMEL-CLIP: Channel-aware multimodal electroencephalography-text alignment for generalizable brain foundation models}

%% Author name(s)
%% Corresponding author address (full address is required)
%% Corresponding author email address
\author[a]{Hanseul Choi} 
\author[a]{Jinyeong Park} 
\author[a]{Seongwon Jin} 
\author[b]{Sungho Park\corref{cor1}} 
\ead{sungho.park@inha.ac.kr}
\author[a,c]{\\ Jibum Kim\corref{cor2}}
\ead{jibumkim@inu.ac.kr}
\cortext[cor1]{Corresponding to: Department of Artificial Intelligence, Inha University, Michuhol-gu, 22212, Incheon, Republic of Korea}
\cortext[cor2]{Corresponding to: Department of Computer Science and Engineering, Incheon National University, Yeonsu-gu, 22012, Incheon, Republic of Korea}

%% Author affiliation
\affiliation[a]{organization={Department of Computer Science and Engineering},%Department and Organization
            addressline={Incheon National University}, 
            city={Yeonsu-gu},
            postcode={22012}, 
            state={Incheon},
            country={Republic of Korea}}

\affiliation[b]{organization={Department of Artificial Intelligence},%Department and Organization
            addressline={Inha University}, 
            city={Michuhol-gu},
            postcode={22212}, 
            state={Incheon},
            country={Republic of Korea}}

\affiliation[c]{organization={Center for Brain-Machine Interface},%Department and Organization
            addressline={Incheon National University}, 
            city={Yeonsu-gu},
            postcode={22012}, 
            state={Incheon},
            country={Republic of Korea}}

\begin{abstract}
Electroencephalography (EEG) foundation models have shown promise for learning generalizable representations, yet they remain sensitive to channel heterogeneity, such as changes in channel composition or ordering. We propose channel-aware multimodal EEG–text alignment contrastive language-image pretraining (CAMEL-CLIP), a contrastive EEG-text multimodal foundation model designed to be robust to heterogeneous channel configurations and widely applicable to diverse downstream tasks. CAMEL-CLIP combines three components: (1) channel attribute-based positional encoding, which identifies channels using their anatomical attributes and reference type; (2) dynamic channel projection, which preserves channel-specific information across varying numbers of EEG channels and combines textual reports with channel information to support channel-wise EEG–text matching; and (3) dual-level contrastive learning, which aligns EEG signals with text at both the individual-channel and whole-sample levels. Experimental results show that CAMEL-CLIP achieves the highest mean balanced accuracy among the evaluated baselines across multiple datasets under linear probing.
CAMEL-CLIP also shows high classification performance when transferred between different channel configurations without classifier retraining.
\end{abstract}

\begin{keyword}
Brain–computer interface \sep Electroencephalography \sep Multimodal learning \sep Foundation model \sep Contrastive learning \sep Channel heterogeneity 
\end{keyword}

\end{frontmatter}

\section{Introduction}
\label{sec1}
Electroencephalography (EEG) is widely used across clinical applications and brain-computer interface (BCI) research, with applications including seizure detection, depression monitoring, Alzheimer’s disease diagnosis, motor imagery, and image reconstruction. To decode EEG signals, numerous deep learning models have been proposed \cite{ay2019automated, shoeibi2021epileptic, solis2024machine}. These approaches can be broadly categorized according to their training strategies and applicability to downstream tasks.

Task-specific models for EEG decoding include EEGNet \cite{lawhern2018eegnet}, EEG Conformer \cite{song2022eeg}, and CNN-LSTM feature fusion network (FFCL) \cite{li2022motor}. These models are lightweight in terms of model parameter size and can effectively process EEG signals. However, since only a limited number of subjects are typically available for each task, they have difficulty in exploiting large-scale datasets. Consequently, they are vulnerable to domain shifts such as inter-subject variability and often exhibit poor generalization performance \cite{saha2020intra}.

To improve generalization beyond individual tasks, self-supervised brain foundation models such as Large Brain Model (LaBraM) \cite{jiang2024large} and a criss-cross brain foundation model for EEG decoding (CBraMod) \cite{wang2024cbramod} have been proposed. By using large-scale datasets, these models are able to learn general-purpose representations and significantly outperform earlier approaches, while showing relatively strong robustness to inter-subject variability. However, they remain sensitive to cross-dataset domain shifts, such as differences in recording settings and changes in channel configurations, and can suffer substantial performance degradation without an additional fine-tuning step \cite{xiong2025eeg}.

Existing models typically learn inter-channel dependencies tied to the channel ordering defined in the training dataset. Consequently, when a domain shift involves changes in channel composition or ordering, the model often fails to adapt effectively \cite{zhang2025brain}. The model cannot effectively reuse its learned knowledge when extracting features under a new channel configuration. This makes it difficult to use pretrained models via linear probing \cite{kuruppu2025eeg}. This problem is referred to as cross-dataset channel heterogeneity. Chen et al. \cite{chen2025hear} proposed an EEG foundation model that leverages the spatial coordinates of electrodes to address this issue. However, electrode coordinates alone do not distinguish reference schemes, which can substantially affect signal characteristics \cite{yao2019which}. Moreover, subject-specific electrode coordinates are not always available in public EEG datasets, whereas channel names and montage information are more commonly provided.

Most existing brain foundation models focus on learning patterns within EEG signals. Extending pretraining to paired EEG and text allows models to learn how these patterns relate to text descriptions of the recordings. To overcome the limitations of unimodal learning, Ndir et al. \cite{ndir2025eeg} proposed the first contrastive language-image pretraining (CLIP)-based multimodal model, named EEG-CLIP. For effective EEG-text alignment, they employed the TUAB dataset, an EEG-text paired dataset. They demonstrated that text-based classification can be performed without an additional classifier. It also highlights the potential of using contrastive learning for EEG-text multimodal models. However, transferring EEG-CLIP to downstream datasets with different channel configurations remains challenging, as its encoder learns representations tied to the channel configuration used during pretraining.

We propose a contrastive EEG-text multimodal foundation model, namely channel-aware multimodal EEG-text alignment contrastive language-image pretraining (CAMEL-CLIP), designed to be robust to changes in channel configurations. First, to mitigate channel heterogeneity, we introduce channel attribute-based positional encoding (CAPE). Inspired by the EEG electrode naming convention, CAPE encodes the region, hemisphere, channel number, and reference type of each channel. This allows the model to identify channels regardless of their ordering in the input, while incorporating reference information that electrode coordinates alone do not represent.

Second, many previous works either assume a fixed number of channels or perform pooling along the channel axis, thereby constraining the model to produce only fixed-length embeddings. To address this limitation, we propose dynamic channel projection (DCP), which combines a clinical report embedding with a channel knowledge embedding to generate a channel-specific text embedding. This constructs a separate text embedding for each EEG channel, enabling alignment with the EEG embedding of that channel across varying channel configurations. Furthermore, to enhance model robustness, we employ channel-wise augmentation techniques such as dynamic channel removal, which removes random channels during training.

Third, we use dual-level contrastive learning (DCL) to align EEG and text at both the channel and sample levels. Channel-level contrastive learning aligns each EEG channel representation with the text representation for the same sample and channel. Sample-level contrastive learning aligns representations formed by concatenating global and channel-level embeddings, thereby incorporating channel-specific information into sample-level alignment.

Finally, the proposed model enables cross-modal tasks beyond the capabilities of unimodal EEG models, such as EEG-text retrieval. Specifically, by retrieving relevant text reports directly from EEG signals, it offers the potential to provide contextual information for clinical interpretation.

\begin{figure}[t]
    \centering
    \begin{minipage}[b]{0.48\linewidth}
    \centering
    \subfloat[EEG-CLIP\label{fig1a}]{
        \includegraphics[width=\linewidth]{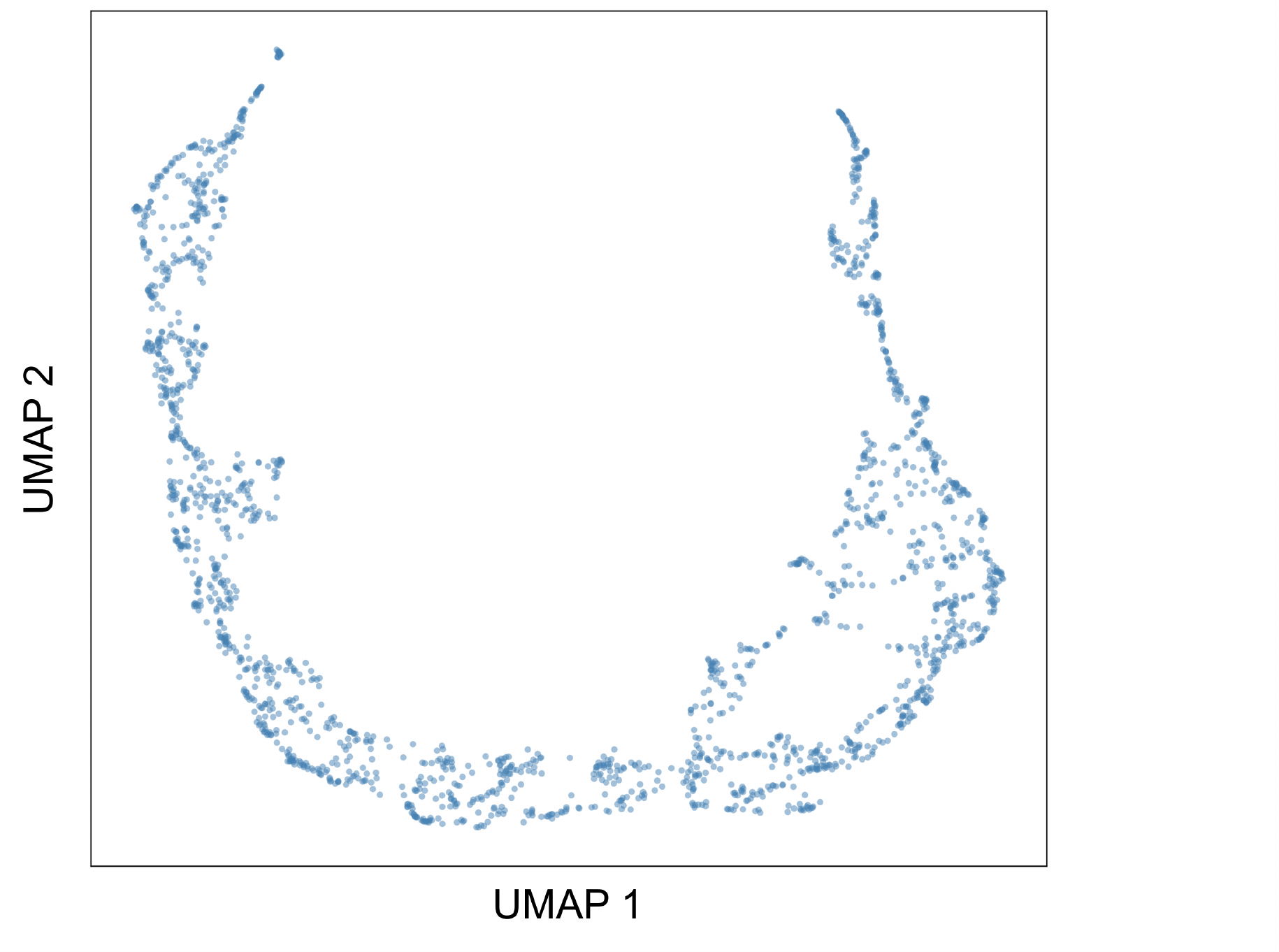}
    }
    \end{minipage}
    \hfill
    \begin{minipage}[b]{0.48\linewidth}
    \centering
    \subfloat[CAMEL-CLIP\label{fig1b}]{
        \includegraphics[width=\linewidth]{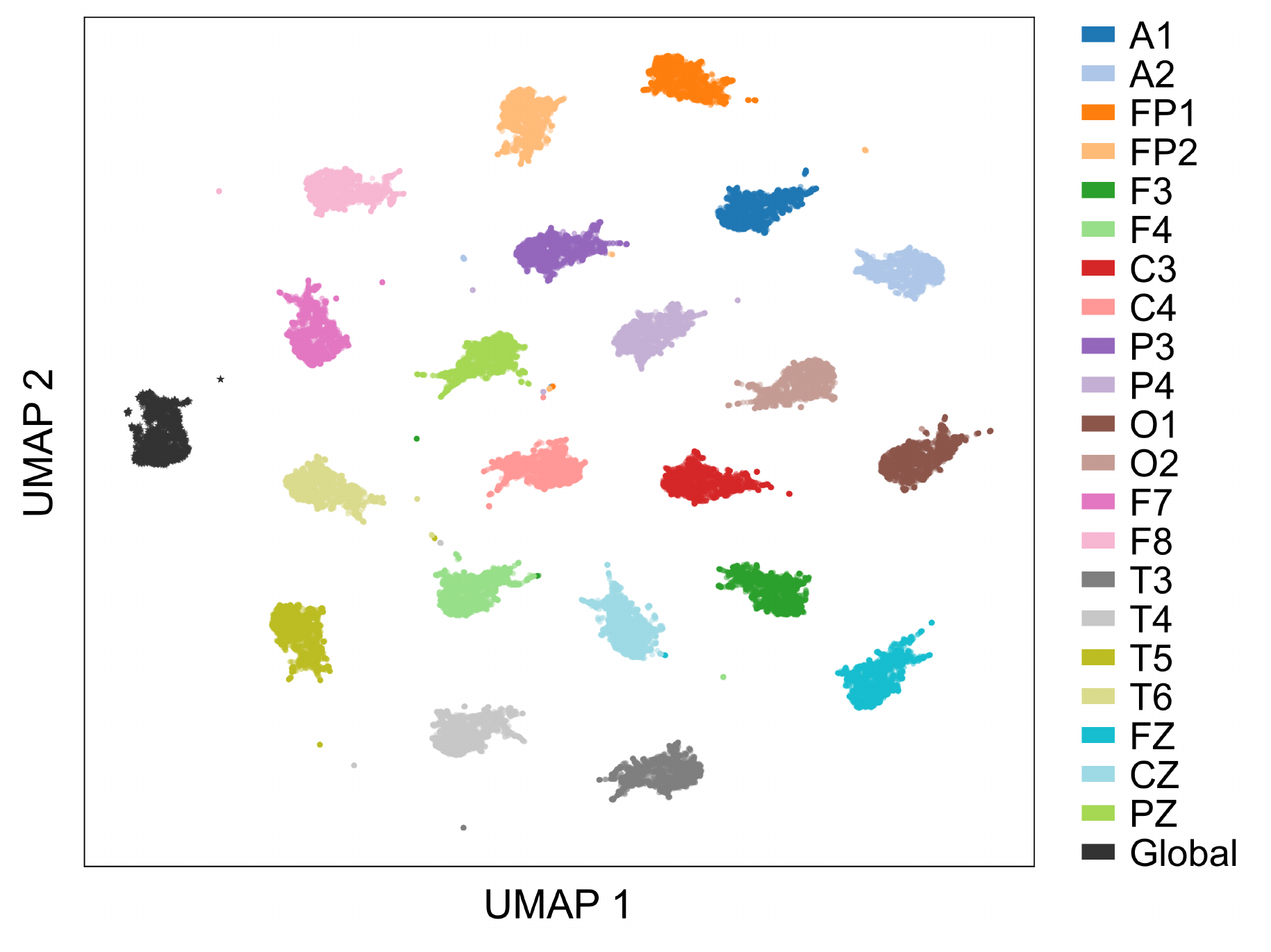}
    }
    \end{minipage}
    \caption{\textbf{UMAP visualization of EEG representations from (a) EEG-CLIP and (b) CAMEL-CLIP.} Each point represents an EEG sample in (a) and a channel-level or global EEG embedding in (b). EEG-CLIP provides a single sample-level representation, shown in blue. In CAMEL-CLIP, colors distinguish the 21 EEG channels and the global embedding, as indicated in the legend. UMAP 1 and UMAP 2 denote the two dimensions of the UMAP projection.}
    \label{fig1}
\end{figure}

Figure \ref{fig1} compares the representation distributions of EEG-CLIP and CAMEL-CLIP using UMAP. EEG-CLIP integrates information from all channels into a single sample-level embedding, so each point in Figure \ref{fig1a} represents an entire EEG sample. These points are not colored by individual channel identity and are instead displayed uniformly in blue. CAMEL-CLIP produces channel-level embeddings and a global embedding. In Figure \ref{fig1b}, different colors denote the 21 channels and the global embedding, as specified in the legend. Points with the same color therefore represent embeddings of the same channel across different EEG samples, or the global embeddings across samples. Embeddings from the same channel form a relatively compact distribution, while the distributions of different channels appear distinct in the UMAP projection. This suggests that the embeddings preserve channel identity across EEG samples while distinguishing between channels.

\begin{figure}[t]
    \centering
    \begin{minipage}[b]{0.31\linewidth}
    \centering
    \subfloat[Previous task-specific models.\label{fig2a}]{
        \includegraphics[width=\linewidth]{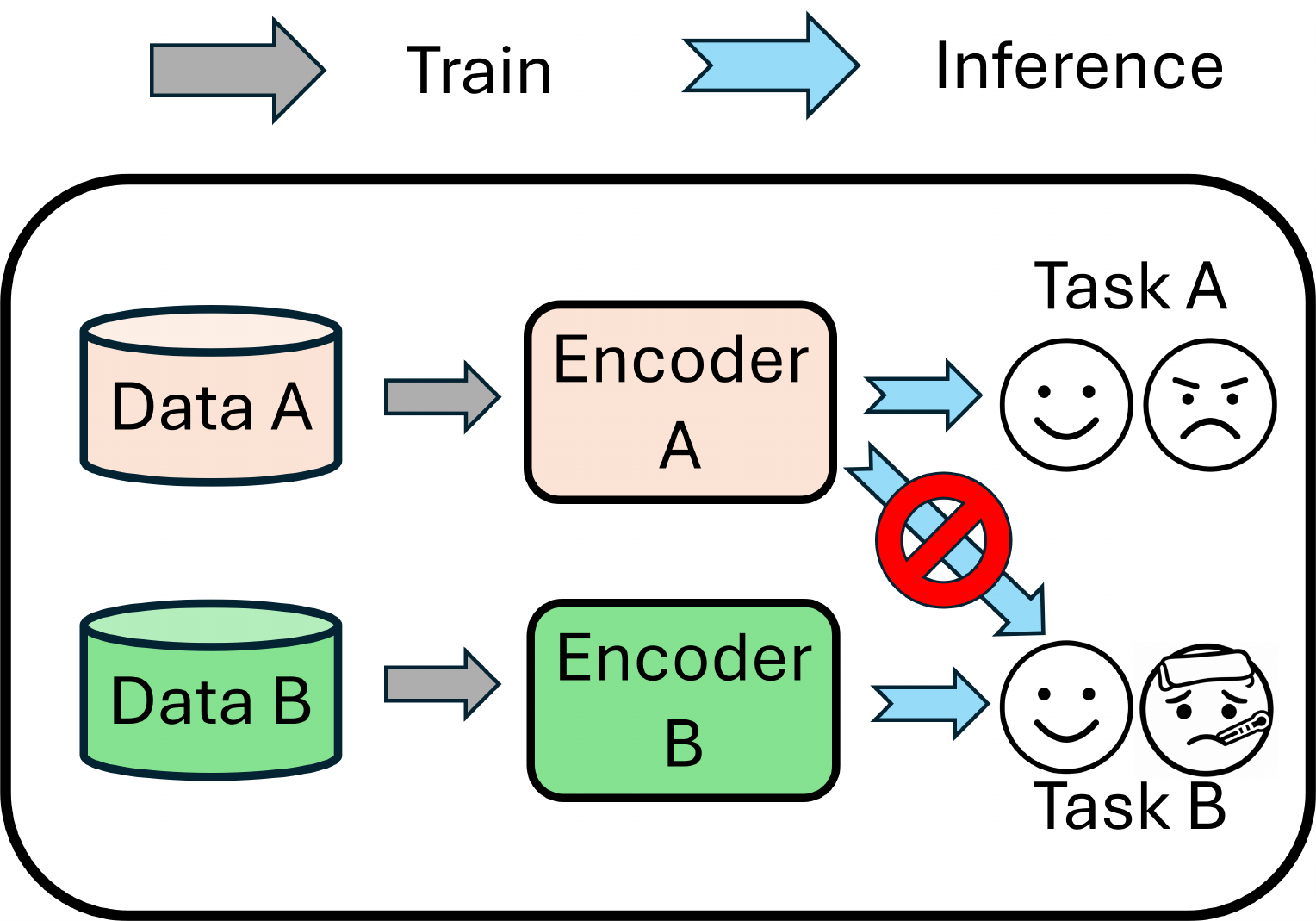}
    }
    \end{minipage}
    \hspace{0.005\linewidth}
    \begin{minipage}[b]{0.31\linewidth}
    \centering
    \subfloat[Previous brain foundation models.\label{fig2b}]{
        \includegraphics[width=\linewidth]{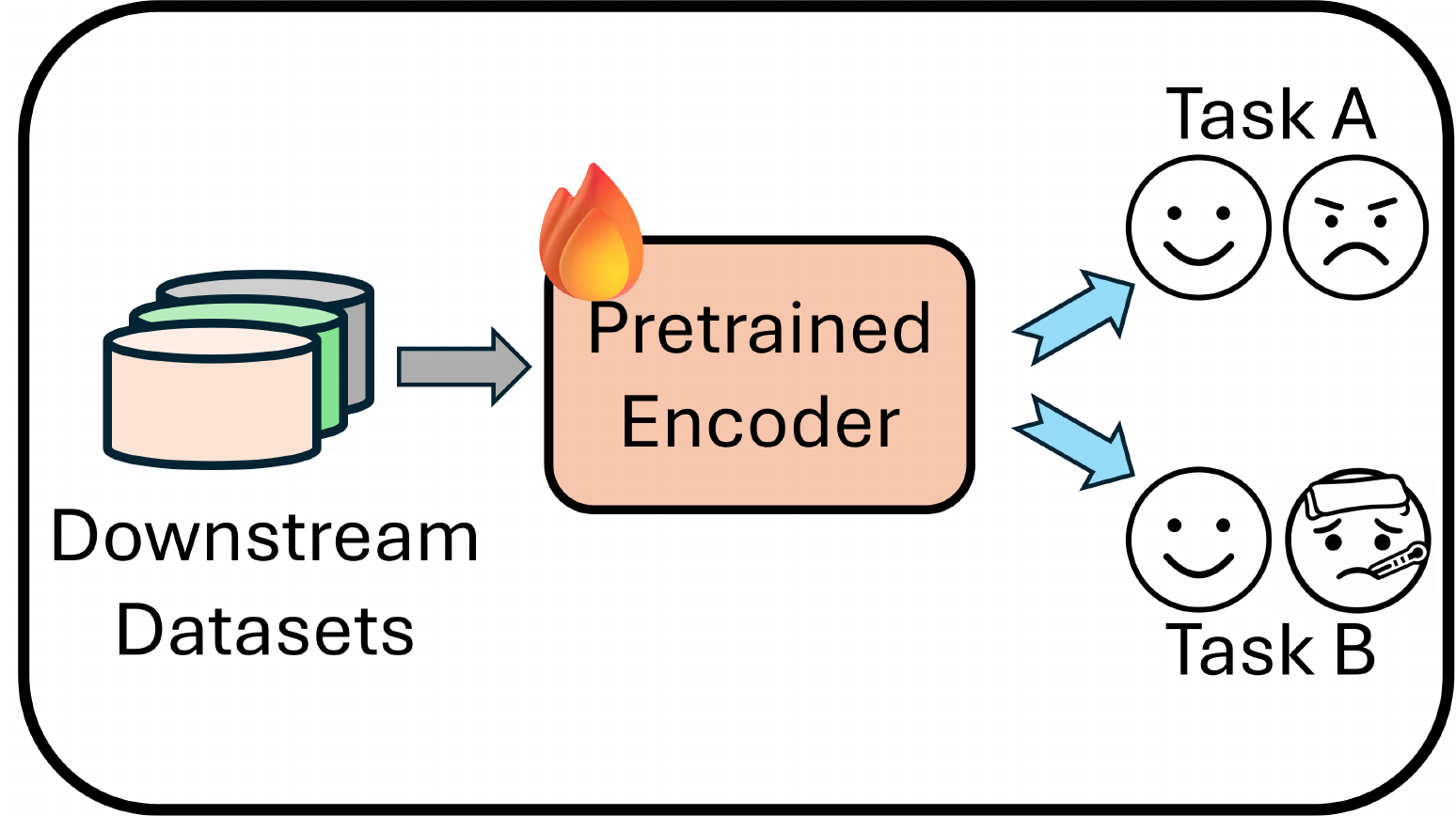}
    }
    \end{minipage}
    \hspace{0.005\linewidth}
    \begin{minipage}[b]{0.31\linewidth}
    \centering
    \subfloat[CAMEL-CLIP.\label{fig2c}]{
        \includegraphics[width=\linewidth]{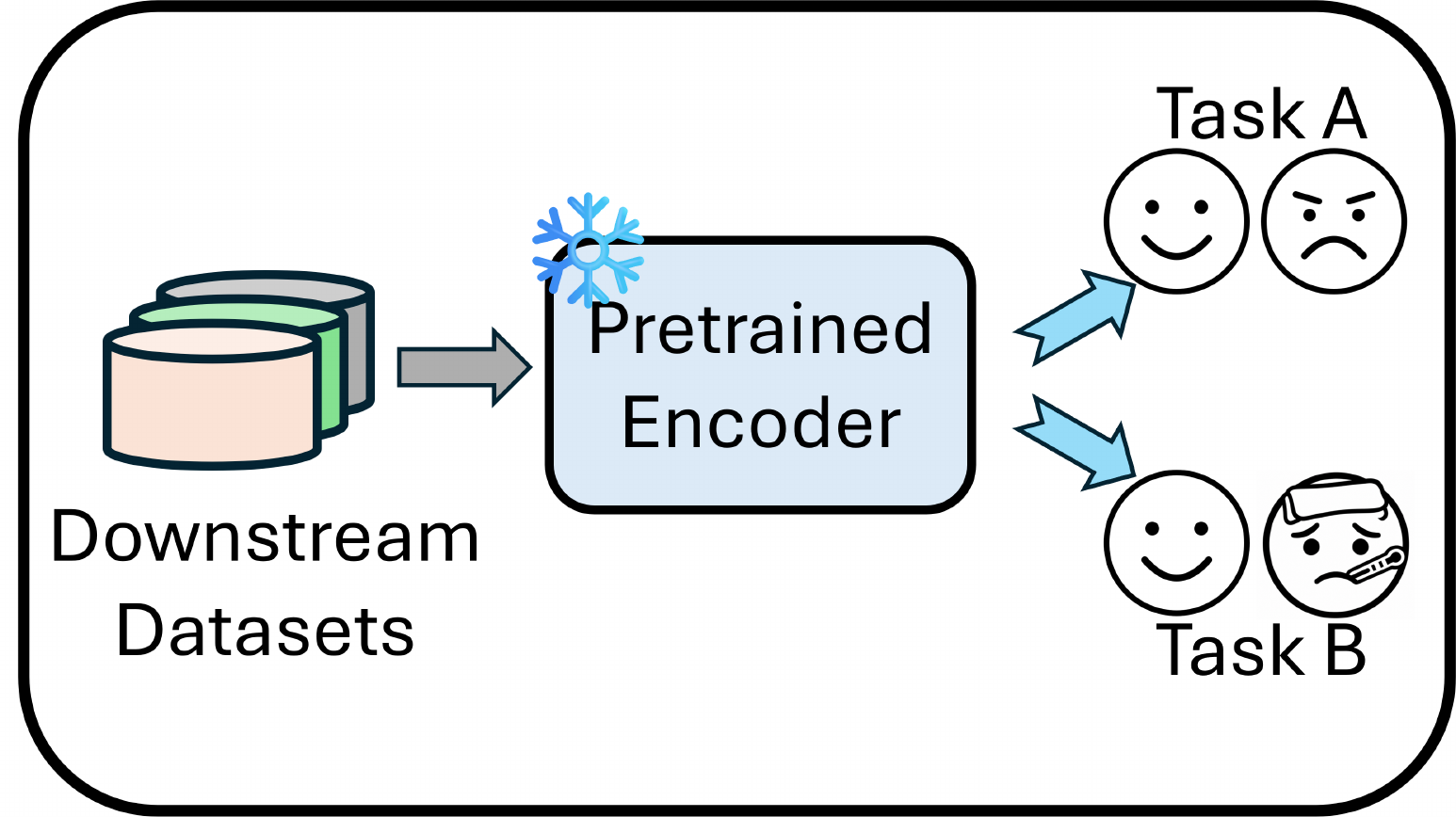}
    }
    \end{minipage}
    \caption{\textbf{Comparison of deep learning-based EEG decoding approaches.} (a) Conventional task-specific models, where a separate EEG encoder is trained for each task and dataset. (b) Prior brain foundation models, which enable multi-task transfer via pretrained weights without necessarily retraining or fine-tuning the entire EEG encoder. (c) CAMEL-CLIP, which mitigates channel heterogeneity and supports multiple tasks with linear probing alone.}
    \label{fig2}
\end{figure}

Figure \ref{fig2} compares deep learning-based EEG decoding approaches. Figure 2(a) illustrates task-specific models, where a separate encoder is trained for each task and dataset \cite{lawhern2018eegnet, song2022eeg}. Figure 2(b) shows the use of pretrained brain foundation models with full fine-tuning on downstream datasets \cite{jiang2024large, wang2024cbramod}. Figure 2(c) presents CAMEL-CLIP, where a single pretrained encoder supports multiple tasks via linear probing, with only a task-specific classifier trained while the encoder remains frozen.

The main contributions of this paper are summarized as follows:
\begin{enumerate}
    \item We propose CAMEL-CLIP, an EEG-text multimodal foundation model designed to be resilient to changes in channel configurations and support diverse downstream tasks through linear probing.
    \item We propose CAPE, DCP, and DCL to represent channel identity, construct channel-specific text representations, and integrate these representations into both channel-level and sample-level EEG--text alignment to address channel heterogeneity.
    \item We demonstrate the cross-modal capabilities of CAMEL-CLIP through EEG--text retrieval, highlighting its potential to provide contextual information for clinical interpretation.
    \item CAMEL-CLIP achieves the highest mean balanced accuracy among the evaluated baselines across multiple datasets under linear probing and shows robustness when transferred between different channel configurations without classifier retraining.
\end{enumerate}

\section{Related works}
\label{sec2}

\subsection{Task-specific EEG decoding models}
\label{sec2.1}

Lawhern et al. \cite{lawhern2018eegnet} proposed EEGNet, a lightweight architecture that effectively extracts spatiotemporal EEG features using depthwise separable convolutions. Subsequently, Li et al. \cite{li2022motor} introduced FFCL, which improves classification performance by fusing features extracted from different layers of convolutional neural network (CNN) and long short-term memory (LSTM) backbones. While these task-specific models are practically efficient due to their small parameter footprint, they struggle to learn generalizable representations from large-scale datasets. Furthermore, they remain highly susceptible to inter-subject variability even within small datasets, often necessitating repeated data collection and model retraining for each novel subject or target task.

\subsection{Brain foundation models}
\label{sec2.2}

Many previous works have proposed brain foundation models for neural decoding. Yang et al. \cite{yang2023biot} developed biosignal transformer (BIOT), which utilizes a standardized tokenization strategy to extract universal features across diverse physiological signals like EEG and ECG. Focusing on architectural refinement, Wang et al. \cite{wang2024cbramod} presented CBraMod. This model leverages a criss-cross attention mechanism to independently process the distinct spatial and temporal dynamics inherent in EEG recordings. Furthermore, Jiang et al. \cite{jiang2024large} pushed the boundaries of scale with LaBraM, training on an extensive 2,500-hour corpus. By implementing a Fourier-based neural tokenizer, they established one of the most extensively parameterized models in the field. However, despite these substantial breakthroughs, a critical gap remains: these state-of-the-art frameworks are exclusively unimodal. This singular focus inherently precludes them from executing complex cross-modal operations, such as mapping EEG signals directly to textual clinical reports.

\subsection{EEG-text multimodal contrastive learning models}
\label{sec2.3}

Recent studies have explored contrastive learning between EEG signals and textual data. Yan et al. \cite{yan2025cross} proposed an EEG-text multimodal contrastive learning model for emotion recognition. By aligning EEG representations to the rich semantic space provided by a text encoder, their method learns domain-invariant representations, and empirical results indicate that multimodal contrastive learning can improve cross-domain generalization. However, due to the scarcity of informative EEG-text paired data, the approach is largely limited to emotion recognition. Ndir et al. \cite{ndir2025eeg} leveraged EEG recordings and clinical text reports to train a model using CLIP-based contrastive learning \cite{radford2021learning}. While this work demonstrated the feasibility of text-based classification in the EEG domain, it reported limited generalization, achieving approximately 0.56 balanced accuracy on a gender classification task. Overall, existing contrastive EEG-text multimodal models are often either task-specific or insufficiently generalizable, limiting their applicability to diverse downstream tasks.

\subsection{Channel-Invariant EEG models}
\label{sec2.4}

To manage channel inconsistencies across datasets, Patil et al. \cite{patil2024coordconformer} utilized 3D positional data to capture the spatial dependencies between electrodes. However, public EEG datasets often lack subject-specific 3D electrode coordinates. Similarly, Chen et al. \cite{chen2025hear} introduced a foundational architecture relying on static electrode positions extracted from a standardized head template. However, coordinate-based encoding may have difficulty capturing differences in reference schemes. Shifting away from spatial coordinates, Yi et al. \cite{yi2023learning} developed a pretraining framework with multi-dimensional position encoding, multi-level channel hierarchy, multi-stage pretraining strategy (MMM) to project varying channel setups onto a static 62-node graph. Nonetheless, this fixed mapping struggles to accommodate channel arrays that fall outside the assumed structure, and extracting differential entropy features risks discarding critical signal information. Döner et al. \cite{doner2025luna} proposed LUNA, which uses learned queries and cross-attention to map multi-channel EEG signals into a fixed-size latent representation, enabling the model to accommodate different electrode configurations.

\section{Methods}
\label{sec3}

\subsection{Overall architecture}
\label{sec3.1}

\begin{figure}[htbp]
    \centering
    \makebox[\linewidth][c]{\includegraphics[width=185mm]{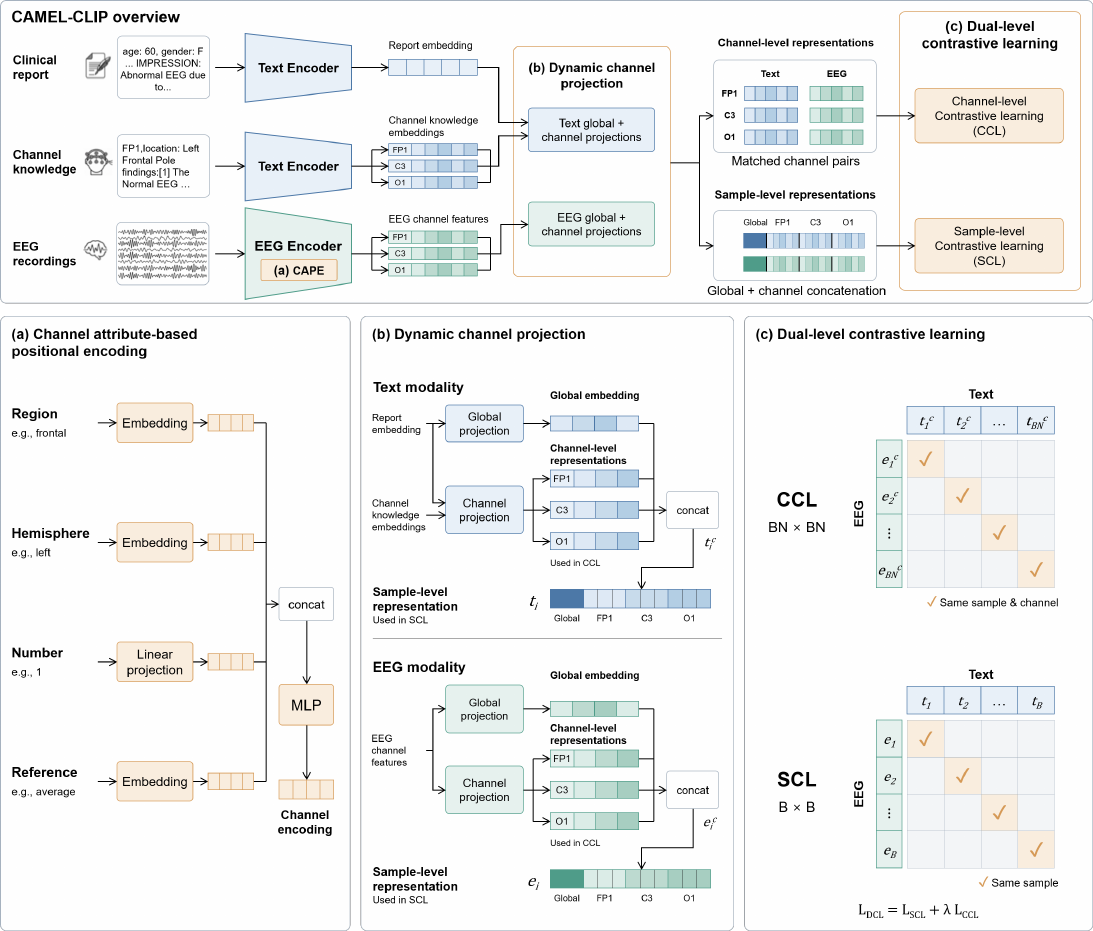}}
    \caption{\textbf{Framework of CAMEL-CLIP.} (a) Channel attribute-based positional encoding. (b) Dynamic channel projection for channel-wise embeddings. (c) Dual-level contrastive learning at the channel and sample levels. The overview and module details distinguish channel-level and sample-level representations and the corresponding contrastive matching objectives.}
    \label{fig3}
\end{figure}

Figure 3 provides an overview of the proposed model, CAMEL-CLIP. The model has five key components: a text encoder, an EEG encoder, channel attribute-based positional encoding (CAPE), dynamic channel projection (DCP), and dual-level contrastive learning (DCL). Given an EEG-text pair as input, the EEG and text encoders extract modality-specific features. The EEG encoder incorporates CAPE to encode channel identity using anatomical attributes and reference type. DCP combines a clinical report embedding with a channel knowledge embedding to generate a channel-specific text embedding. DCL then aligns EEG and text representations at both the channel and sample levels, learning correspondence between EEG and text for each channel and incorporating channel-specific information into sample-level alignment.

\paragraph{\textbf{Text encoder}} We adapt ClinicalBERT \cite{huang2019clinicalbert} as the text encoder in our model. ClinicalBERT is a BERT \cite{devlin2019bert} model pretrained on electronic health records (EHRs) and is well aligned with the linguistic characteristics of clinical EEG reports, which differ substantially from general-domain text in both style and context. During contrastive training, we initialize the text encoder with pretrained weights and keep it frozen to mitigate overfitting.

\paragraph{\textbf{EEG encoder}} We adapt the self-supervised pretrained model CBraMod \cite{wang2024cbramod} as the EEG encoder. CBraMod is designed to disentangle temporal and spatial characteristics during representation learning. To generate channel positional encodings independent of channel order, we introduce a new CAPE module. Aside from this modification, we initialize the EEG encoder with the pretrained CBraMod weights and fine-tune the remaining parameters during training.

\subsection{Channel attribute-based positional encoding}
\label{sec3.2}

To alleviate the performance drop induced by channel heterogeneity \cite{xiong2025eeg}, we introduce CAPE. Drawing inspiration from established EEG nomenclatures \cite{jasper1958ten}, CAPE categorizes channels using four distinct features: (1) region (scalp area), (2) hemisphere (lateral position), (3) number (proximity to the midline), and (4) reference type. This attribute-based approach represents both anatomical identity and reference information, whereas electrode coordinates alone do not represent differences in reference schemes \cite{yao2019which}. CAPE uses readily available channel names and montage information, allowing it to represent channels without requiring subject-specific 3D electrode coordinates. For the encoding process, categorical variables (\textit{i.e.,} region, hemisphere, and reference type) are mapped via learnable embedding layers, while the channel number is treated as a numerical input and mapped through a linear projection. These four intermediate representations are subsequently concatenated and fed into a multilayer perceptron, yielding a channel embedding that incorporates anatomical and reference attributes.

Bipolar montages measure the potential difference between a pair of electrodes \cite{yao2019which}. In CAPE, we define the embedding of a bipolar channel as the mean of its constituent electrode embeddings. This design enables bipolar channels to be represented using the same electrode embeddings as unipolar channels, supporting the transfer of learned electrode representations across montage types.

\subsection{Dynamic channel projection}
\label{sec3.3}

To preserve the distinct representational properties of individual EEG channels, we propose DCP, which constructs channel-specific text embeddings corresponding to individual EEG channels. Specifically, channel-specific text embeddings are produced by fusing a patient’s clinical report embedding with a channel knowledge embedding via a fully connected layer. These channel knowledge embeddings are derived by encoding anatomical and medical descriptions sourced from literature using the text encoder. By constructing a corresponding text embedding for each available EEG channel, DCP supports channel-wise EEG--text alignment across varying channel counts while retaining channel-specific information.

Furthermore, DCP supports dynamic channel removal, a stochastic data augmentation strategy applied during training. For any given mini-batch, this technique is triggered with a predefined probability, $p_{remove}$. Once activated, the number of channels to remove is sampled uniformly from the range $[1, C \cdot r_{max}]$, where $C$ is the total channel count and $r_{max}$ specifies the maximum proportion of channels to remove. The selected number of channels is then randomly discarded. For example, given a 20-channel setup with $p_{remove} = 0.3$ and $r_{max} = 0.5$, roughly 30\% of the training batches will randomly lose between 1 and 10 channels. This augmentation exposes the model to different channel configurations during training, encouraging robustness to changes in channel availability.

\subsection{Dual-level contrastive learning}
\label{sec3.4}

We propose DCL to enable the model to extract meaningful features for unseen channels by leveraging learned representations. This training scheme encourages the EEG encoder to jointly learn channel-specific embeddings that capture distinct channel characteristics and a global embedding that summarizes the overall signal. DCL comprises two contrastive objectives, Channel-Level Contrastive Learning (CCL) and Sample-Level Contrastive Learning (SCL), both based on the InfoNCE objective \cite{oord2018representation}.

CCL is introduced to learn channel-level representations. In CCL, the channel-wise EEG embedding and the corresponding channel-wise text embedding from the same sample and the same channel are treated as a positive pair. All other channel embeddings from the same sample (\textit{i.e.}, different channels) as well as embeddings from other samples are treated as negative pairs. For training efficiency, we randomly sample $N$ channels from each sample in a mini-batch and apply contrastive learning at the channel-embedding level. 
The CCL loss is defined as follows:

\begin{equation}
\resizebox{0.92\textwidth}{!}{
$L_{CCL} = -\frac{1}{2BN} \sum_{i=1}^{BN} \left( \log \frac{\exp(e_{i}^{c \top} t_{i}^{c} / \tau)}{\sum_{j=1}^{BN} \exp(e_{i}^{c \top} t_{j}^{c} / \tau)} + \log \frac{\exp(t_{i}^{c \top} e_{i}^{c} / \tau)}{\sum_{j=1}^{BN} \exp(t_{i}^{c \top} e_{j}^{c} / \tau)} \right)$
},
\end{equation}
where $e_i^c$ denotes the $i$-th channel-wise EEG embedding and $t_i^c$ denotes the $i$-th channel-wise text embedding. $B$ is the batch size, $N$ is the number of sampled channels per sample, and $\tau$ is the temperature parameter.

To augment channel-specific representations with information shared across channels, we construct a global EEG embedding by applying attention pooling over the channel embeddings and their corresponding channel features. The corresponding global text embedding is obtained by projecting only the clinical report embedding, without incorporating channel information. For each modality, the global embedding is concatenated with the channel-level embeddings to form a sample-level representation. SCL aligns these concatenated EEG and text representations, treating representations from the same sample as positive pairs and those from different samples as negative pairs, thereby incorporating channel-specific information into sample-level alignment. The SCL loss is defined as follows:

\begin{equation}
L_{SCL} = -\frac{1}{2B} \sum_{i=1}^{B} \left( \log \frac{\exp(e_{i}^{\top} t_{i} / \tau)}{\sum_{j=1}^{B} \exp(e_{i}^{\top} t_{j} / \tau)} + \log \frac{\exp(t_{i}^{\top} e_{i} / \tau)}{\sum_{j=1}^{B} \exp(t_{i}^{\top} e_{j} / \tau)} \right),
\end{equation}
Here, $e_i$ and $t_i$ denote the concatenated sample-level EEG and text representations for the $i$-th sample, respectively. $B$ is the batch size, and $\tau$ is the temperature parameter.

To prevent the model from being overly dominated by the CCL loss such that the intended SCL objective is not effectively optimized, we weight CCL by a factor of $\lambda$ when combining the losses. The overall objective is defined as follows:
\begin{equation}
L_{DCL} = L_{SCL} + \lambda \cdot L_{CCL}.
\end{equation}

\subsection{Synthetic Report Ensemble Prompt Generation}
\label{sec3.5}

\begin{figure}[htbp]
    \centering
    \begin{minipage}[b]{0.9\linewidth}
    \centering
    \subfloat[Proposed text prompt generation method.\label{fig4a}]{
        \includegraphics[width=\linewidth]{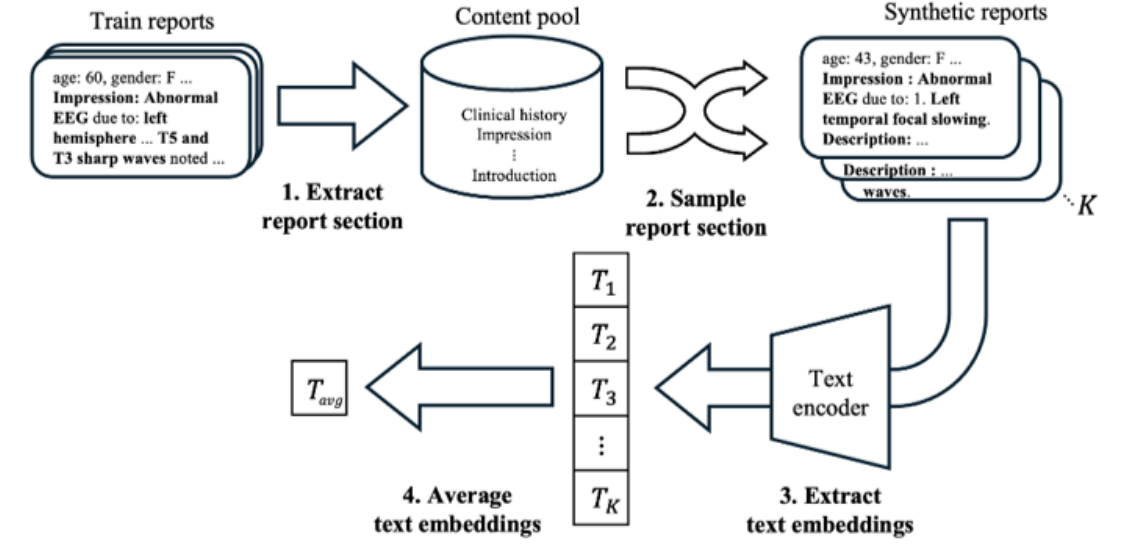}
    }
    \end{minipage}
    
    \vspace{0.8em}
    \begin{minipage}[b]{0.9\linewidth}
    \centering
    \subfloat[An example of generated synthetic report.\label{fig4b}]{
        \includegraphics[width=\linewidth]{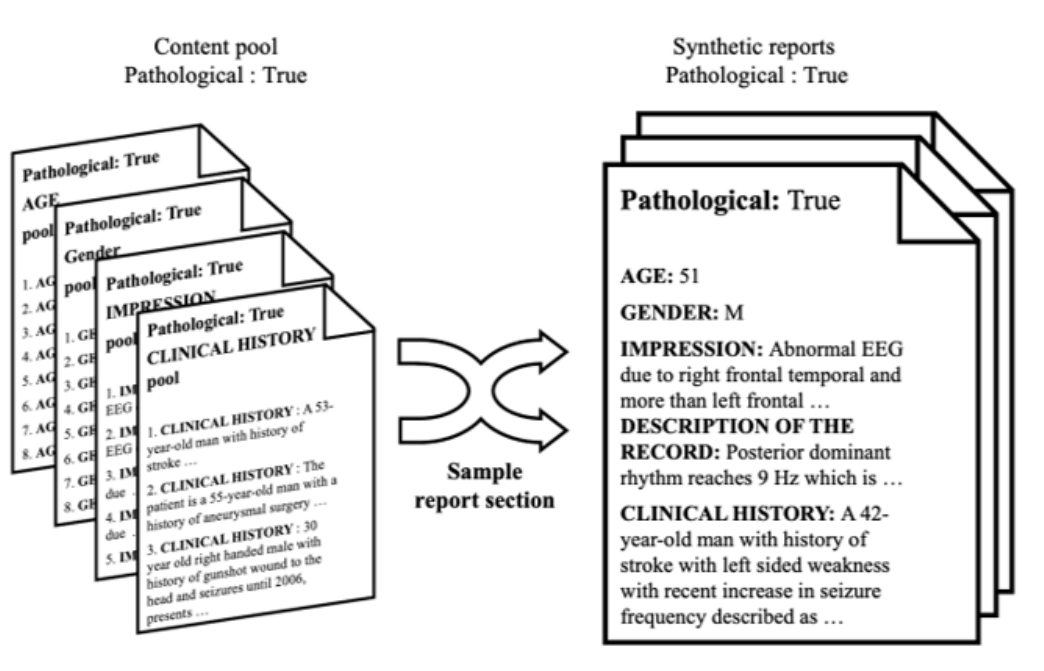}
    }
    \end{minipage}
    \caption{\textbf{Pipeline of proposed synthetic report ensemble prompt generation.} (a) The proposed text prompt generation method for text-based classification generates multiple synthetic reports and averages their embeddings. (b) An example of generated synthetic report.}
    \label{fig4}
\end{figure}

Text embeddings derived from conventional single-sentence prompts often exhibit a significant distributional shift from the representations inherently learned by the model. Consequently, these simplistic prompts fail to accurately capture the true semantic space of each label. To resolve this discrepancy, we introduce a novel technique termed synthetic report ensemble prompt generation. This approach first dissects training reports into discrete structural sections, aggregating them by class to form label-specific content pools. During inference, we synthesize $K$ distinct reports by uniformly sampling and concatenating one text segment from each section's pool. The final text prototype for a given label is then computed as the average embedding of these $K$ synthetic reports. For instance, in a binary pathology classification task (\textit{normal} vs. \textit{abnormal}), the \textit{normal} prototype is formulated by independently drawing segments from pools such as clinical history, medications, and impression—all exclusively sourced from normal cases. After generating $K$ such reports, their embeddings are averaged to establish the robust \textit{normal} prototype.
 
Figure 4 (a) illustrates the proposed framework for generating averaged report embeddings. Figure 4 (b) shows an example of generated synthetic report. The label embedding for label $y$ is computed by averaging the text-encoder embeddings of $K$ synthetic reports:

\begin{equation}
E_y = \frac{1}{K} \sum_{k=1}^{K} \text{Textencoder}(T_{synthetic, y}^{k}) ,
\end{equation}
where $K$ is the number of synthetic reports and $T_{\mathrm{synthetic}, y}^{k}$ denotes the $k$-th synthetic report for label $y$.

\begin{figure}[t]
    \centering
    \includegraphics[width=0.8\linewidth]{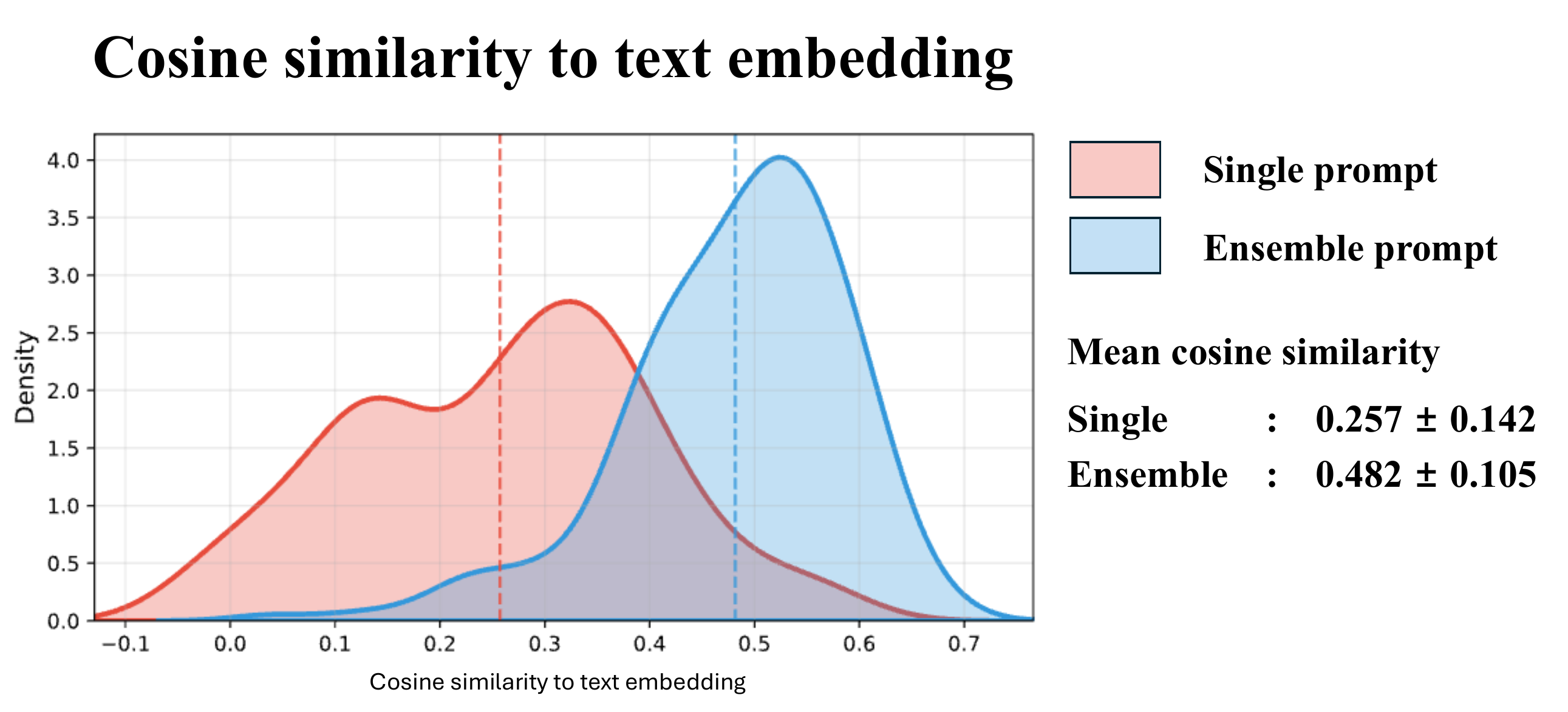}
      \caption{\textbf{Cosine similarity distributions across prompting strategies.} Cosine similarity between prompts for normal labels in the pathological task and text embeddings from the validation set. The prompts from the proposed ensemble method show a higher mean cosine similarity.}
    \label{fig5}
\end{figure}

To demonstrate the efficacy of our averaged report embedding strategy, we evaluated the cosine similarity distributions on the TUAB dataset \cite{obeid2016temple}. Specifically, we measured the alignment between the text prototypes generated for the \textit{normal} pathology label and the actual distribution of clinical reports in the validation set. As illustrated in Figure 5, the embeddings derived via our proposed ensemble approach exhibit a substantially higher mean cosine similarity than the single-prompt baseline. Quantitatively, the single-prompt method yields a mean similarity of 0.257, whereas our proposed technique achieves 0.482. These findings confirm that the synthetic report ensemble prompt generation method captures the semantic distribution of the target labels far more accurately than conventional single-sentence prompts.

\section{Experimental settings}
\label{sec4}

\subsection{Datasets}
\label{sec4.1}

\paragraph{\textbf{TUAB dataset}} 
This dataset is a subset of the Temple University Hospital EEG Data Corpus \cite{obeid2016temple} and labels each patient as normal or abnormal. The recordings use an average reference montage with 21 channels. Following the data split used in EEG-CLIP \cite{ndir2025eeg}, the TUAB dataset includes 1,237 normal and 893 abnormal subjects in the training split, and 148 normal and 105 abnormal subjects in the evaluation split. Each patient report consists of 15 fields, including metadata such as \textit{age} and \textit{gender}, as well as clinical symptoms and examination findings. In this work, we use the report content to perform binary classification on three labels: \textit{pathological, gender, and age.} Following the age classification protocol of EEG-CLIP \cite{ndir2025eeg}, we classify patients into two groups: those aged 50 years or older and those younger than 50 years.

\paragraph{\textbf{CHB-MIT dataset}} 
This dataset contains EEG recordings from 22 pediatric patients with intractable seizures \cite{shoeb2009application, goldberger2000physiobank, guttag2010chb}. The recordings use a bipolar montage with 23 channels. Patients 1 to 19 are used for training, patients 20 and 21 for validation, and patients 22 and 23 for evaluation. Following CBraMod \cite{wang2024cbramod}, all EEG segments from the same subject were included in a single split to prevent subject overlap between the training, validation, and test sets.

\paragraph{\textbf{MUMTAZ-2016 dataset}} 
This dataset was collected from 34 patients with major depressive disorder and 30 healthy controls \cite{mumtaz2016mdd}. The recordings use a linked-ears (LE) reference montage with 19 channels. The training split includes 24 patients with major depressive disorder and 19 controls, the validation split includes 5 patients and 4 controls, and the evaluation split includes 5 patients and 5 controls. Following CBraMod \cite{wang2024cbramod}, all EEG segments from the same subject were included in a single split to prevent subject overlap between the training, validation, and test sets.

\subsection{Implementation details}
\label{sec4.2}
We trained the proposed model for 100 epochs with a mini-batch size of 512. Without early stopping, we evaluated text-based classification on the TUAB validation set at every epoch for three tasks (\textit{i.e.}, pathological, gender, and age) and selected the checkpoint with the highest mean balanced accuracy across tasks. Both the global and channel embedding dimensions were set to 64. Dynamic channel removal was applied with $p_{remove}$ = 0.5, $r_{max}$ = 0.8. The $p_{remove}$ was chosen via the hyperparameter search in Table \ref{tab:hyp_CRP}. For CCL, we set the channel sampling number $n$ to 3 based on the hyperparameter search in Table \ref{tab:hyp_CSn}. We set $\lambda$ to 0.5 based on the highest mean balanced accuracy across the three TUAB validation tasks in the grid search reported in Table \ref{tab:lambda}. For evaluation with a classifier, each model was trained five times with five different random seeds. The validation sets were used for model selection and hyperparameter evaluation.

\begin{table}[p]
\small
\setlength{\tabcolsep}{6pt}
\setlength{\extrarowheight}{0pt}
\renewcommand{\arraystretch}{1.3}
\captionsetup{skip=8pt}
\centering
\caption{\textbf{Average text-based classification balanced accuracy across varying dynamic channel removal probability $p_{remove}$.} This table reports the mean balanced accuracy of text-based classification on the TUAB validation set for different removal probabilities.}
\label{tab:hyp_CRP}
\begin{tabular}{@{}lc@{}}
\hline
$\boldsymbol{p}_{\mathbf{remove}}$& \textbf{Balanced accuracy} \\ \hline
0& \tablenum[table-format=1.4]{0.7330}\\ 
0.25& \tablenum[table-format=1.4]{0.7325}\\ 
0.50& \textbf{\tablenum[table-format=1.4]{0.7442}}\\ 
0.75& \tablenum[table-format=1.4]{0.7230}\\\hline
\end{tabular}
\vspace{1.5em}
\centering
\caption{\textbf{Average text-based classification balanced accuracy across CCL sampling values of $\boldsymbol{n}$.} This table reports the mean balanced accuracy of text-based classification on the TUAB validation set for different CCL sampling values of $n$.}
\label{tab:hyp_CSn}
\begin{tabular}{@{}lc@{}}
\hline
$\boldsymbol{n}$& \textbf{Balanced accuracy} \\ \hline
1& \tablenum[table-format=1.4]{0.7387}\\ 
3& \textbf{\tablenum[table-format=1.4]{0.7442}}\\ 
5& \tablenum[table-format=1.4]{0.7416}\\\hline
\end{tabular}
\vspace{1.5em}
\centering
\caption{\textbf{Validation performance across values of $\lambda$.} Mean balanced accuracy is reported for each value of the loss-weighting coefficient.}
\label{tab:lambda}
\begin{tabular}{@{}lc@{}}
\hline
$\boldsymbol{\lambda}$ & \textbf{Balanced accuracy} \\
\hline
0 & \tablenum[table-format=1.4]{0.7268} \\
0.2 & \tablenum[table-format=1.4]{0.7402} \\
0.4 & \tablenum[table-format=1.4]{0.7402} \\
0.5 & \textbf{\tablenum[table-format=1.4]{0.7442}} \\
0.6 & \tablenum[table-format=1.4]{0.7411} \\
0.8 & \tablenum[table-format=1.4]{0.7356} \\
1.0 & \tablenum[table-format=1.4]{0.7342} \\
\hline
\end{tabular}
\end{table}

% Response Table A16

\FloatBarrier

\section{Experiments}
\label{sec5}

In this section, we conduct a diverse set of experiments to evaluate the effectiveness of CAMEL-CLIP. Table \ref{tab:exp_summary} summarizes the objectives and procedures of the experiments in this work.

\textbf{Experiment 1} performs text-based classification with single and ensemble prompts to assess the contribution of synthetic report ensemble prompting to text-based classification performance. \textbf{Experiment 2} investigates whether the EEG encoder learned by the proposed framework extracts generalizable features across datasets. We attach an additional classifier and compare performance with task-specific models and EEG foundation models under both full fine-tuning and linear probing settings. \textbf{Experiment 3} conducts an ablation study to evaluate the contribution of each component of the proposed method and compares positional encoding variants to assess the effectiveness of CAPE. \textbf{Experiment 4} evaluates the impact of the proposed CCL by visualizing and quantitatively analyzing channel-wise representations. We also assess robustness to channel heterogeneity by transferring a classifier between different channel configurations without retraining. \textbf{Experiment 5} examines the contribution of text information and the quality of EEG--text alignment through EEG-only versus EEG--text pretraining comparisons, cosine similarity analysis of positive and negative EEG--text pairs, and EEG--text retrieval. \textbf{Experiment 6} evaluates the computational overhead of proposed methods by comparing training cost and inference efficiency with and without DCP and DCL.

% Response Table A6
\begin{table}[p]
\centering
\small
\setlength{\tabcolsep}{4pt}
\renewcommand{\arraystretch}{1.35}
\caption{\textbf{Summary of experiments.} This table summarizes the objective and procedure of each experiment.}
\label{tab:exp_summary}
\makebox[\textwidth][c]{%
\begin{tabular}{@{}>{\raggedright\arraybackslash}p{2.65cm}>{\raggedright\arraybackslash}p{4.4cm}>{\raggedright\arraybackslash}p{2.8cm}>{\raggedright\arraybackslash}p{6.0cm}@{}}
\hline
\textbf{Experiment} & \textbf{Experiments type} & \textbf{Dataset} & \textbf{Goal} \\
\hline
Experiment 1 & Text-based classification & TUAB & Evaluate the performance improvement provided by ensemble prompting \\[0.8em]
\cline{1-4}
\multirow[t]{3}{=}{Experiment 2} & \multirow[t]{3}{=}{Classification with classifier and additional baseline comparisons} & TUAB & \multirow[t]{3}{=}{Evaluate downstream classification performance and compare with competitive EEG models across datasets} \\
\cline{3-3}
 &  & CHB-MIT &  \\
\cline{3-3}
 &  & MUMTAZ-2016 &  \\[0.8em]
\cline{1-4}
\multirow[t]{4}{=}{Experiment 3} & Text-based classification & TUAB & \multirow[t]{4}{=}{Assess the contributions and interactions of CAPE and Dual-level Alignment, and compare positional encoding methods} \\
\cline{2-3}
 & \multirow[t]{2}{=}{Classification with classifier} & CHB-MIT &  \\
\cline{3-3}
 &  & MUMTAZ-2016 &  \\
\cline{2-3}
 & Positional encoding comparison & \begin{tabular}[t]{@{}l@{}}CHB-MIT\\\hline MUMTAZ-2016\end{tabular} &  \\[0.8em]
\cline{1-4}
\multirow[t]{4}{=}{Experiment 4} & \multirow[t]{2}{=}{Channel representation analysis} & TUAB & \multirow[t]{4}{=}{Evaluate channel representation quality and robustness to channel heterogeneity, and assess the effectiveness of channel-level contrastive learning} \\
\cline{3-3}
 &  & CHB-MIT &  \\
\cline{2-3}
 & CCL representation analysis & TUAB &  \\
\cline{2-3}
 & Asymmetric channel mismatch & MUMTAZ-2016 &  \\[0.8em]
\cline{1-4}
\multirow[t]{3}{=}{Experiment 5} & EEG-only vs. EEG-text comparison & \begin{tabular}[t]{@{}l@{}}TUAB\\\hline CHB-MIT\\\hline MUMTAZ-2016\end{tabular} & \multirow[t]{3}{=}{Evaluate EEG-text alignment quality and assess the contribution of the text modality to the performance improvement of CAMEL-CLIP} \\
\cline{2-3}
 & Cosine similarity of positive and negative EEG--text pairs & TUAB &  \\
\cline{2-3}
 & EEG-text retrieval & TUAB &  \\[0.8em]
\cline{1-4}
Experiment 6 & Computational overhead analysis & - & Evaluate the additional computational cost of Dual-level Alignment (DLA) \\
\hline
\end{tabular}}
\end{table}

When using the TUAB dataset, we set the baseline model as the EEG-CLIP \cite{ndir2025eeg}. This is because the TUAB dataset is a multimodal EEG-text dataset and the only existing EEG-text multimodal model based on the TUAB dataset is the EEG-CLIP model.

\FloatBarrier
\subsection{Text-based classification}
\label{sec5.1}

In Experiment 1, we evaluate the contribution of synthetic report ensemble prompting to text-based classification performance on the TUAB dataset. By employing cosine similarity to match EEG embeddings against class-specific text prompts, we perform classification without training an additional classifier. We evaluate both EEG-CLIP and CAMEL-CLIP under single-prompt and ensemble-prompt settings. Comparing the two prompt settings within each model assesses the effect of ensemble prompting, while comparing the models under the same prompt setting assesses their relative classification performance.

Table \ref{tab:tbc_single} shows that ensemble prompting improves balanced accuracy across all three tasks for both models. For EEG-CLIP, balanced accuracy increases from 0.4467 to 0.7905 for pathological classification, from 0.5360 to 0.5608 for gender classification, and from 0.4395 to 0.6718 for age classification. For CAMEL-CLIP, the corresponding improvements are from 0.5640 to 0.8158, from 0.6437 to 0.6463, and from 0.6751 to 0.7125. CAMEL-CLIP also achieves higher balanced accuracy than EEG-CLIP on all three tasks under both prompt settings. These results show that ensemble prompting benefits both models, while CAMEL-CLIP retains its performance advantage under matched prompt settings.

% Response Table A15
\begin{table}[!t]
\centering
\small
\setlength{\tabcolsep}{5pt}
\renewcommand{\arraystretch}{1.35}
\caption{\textbf{Text-based classification with single and ensemble prompts.} This table reports text-based classification performance on the three TUAB tasks using a single prompt or ensemble prompts. Bold performance values indicate the highest value for each task across the model and prompt combinations.}
\label{tab:tbc_single}
\label{tab:tbc_ensemble}
\makebox[\textwidth][c]{%
\begin{tabular}{@{}lcccc@{}}
\hline
\textbf{Model} & \textbf{Prompt} & \textbf{Pathological} & \textbf{Gender} & \textbf{Age} \\
\hline
\multirow{2}{*}{EEG-CLIP \cite{ndir2025eeg}} & Single & \tablenum[table-format=1.4]{0.4467} & \tablenum[table-format=1.4]{0.5360} & \tablenum[table-format=1.4]{0.4395} \\
 & Ensemble & \tablenum[table-format=1.4]{0.7905} & \tablenum[table-format=1.4]{0.5608} & \tablenum[table-format=1.4]{0.6718} \\

\hline
\multirow{2}{*}{\textbf{CAMEL-CLIP (Ours)}} & Single & \tablenum[table-format=1.4]{0.5640} & \tablenum[table-format=1.4]{0.6437} & \tablenum[table-format=1.4]{0.6751} \\
 & Ensemble & \textbf{\tablenum[table-format=1.4]{0.8158}} & \textbf{\tablenum[table-format=1.4]{0.6463}} & \textbf{\tablenum[table-format=1.4]{0.7125}} \\
\hline
\end{tabular}}
\end{table}

\FloatBarrier
\subsection{Finetuning classification performance}
\label{sec5.3}

In Experiment 2, we investigate the cross-dataset generalization capabilities of the proposed framework. To achieve this, we append a classifier head to our pretrained EEG encoder and evaluate its transferability across multiple datasets. To assess generalization and resilience to channel variations, we primarily focus on the linear probing setting, wherein the EEG encoder remains frozen and only the classifier is optimized. Additionally, for comparison with prior baseline models, we also report outcomes under a full fine-tuning paradigm.

First, we detail the linear probing results on the TUAB dataset using EEG-CLIP as the baseline. As demonstrated in Table \ref{tab:cls_TUAB}, our model achieves higher mean balanced accuracy across all three tasks, with improvements of 2.09, 9.50, and 5.11 percentage points for pathological, gender, and age classification, respectively. All three differences are statistically significant under a two-sided Welch's t-test (p < 0.05).

Second, we examine downstream generalization using the CHB-MIT and MUMTAZ-2016 datasets. The comparisons in Table \ref{tab:cls_CHB} include CBraMod, MMM, and LUNA-Base/Huge alongside the existing baselines. On the CHB-MIT dataset, CAMEL-CLIP achieves the highest mean balanced accuracy in both full fine-tuning and linear probing settings. Under linear probing, its balanced accuracy is significantly higher than that of four of the five baselines after Holm correction.

% Response Table A13
\begin{table}[!t]
\centering
\small
\setlength{\tabcolsep}{5pt}
\renewcommand{\arraystretch}{1.35}
\caption{\textbf{TUAB dataset classification with a trained classifier head.} This table reports linear probing performance on the three TUAB tasks. Values are mean $\pm$ SD. An asterisk (*) indicates a significant difference from CAMEL-CLIP (two-sided Welch's t-test, n = 5 per model, p < 0.05). Each task-specific balanced accuracy comparison constitutes a separate family with one comparison.}
\label{tab:cls_TUAB}
\makebox[\textwidth][c]{%
\begin{tabular}{@{}lccc@{}}
\hline
\textbf{Method} & \textbf{Pathological} & \textbf{Gender} & \textbf{Age} \\
\hline
EEG-CLIP \cite{ndir2025eeg} & \makebox[34pt][c]{\num{0.8148}}\makebox[14pt][c]{$\pm$}\makebox[34pt][c]{\num{0.0027}\rlap{*}} & \makebox[34pt][c]{\num{0.5798}}\makebox[14pt][c]{$\pm$}\makebox[34pt][c]{\num{0.0019}\rlap{*}} & \makebox[34pt][c]{\num{0.6837}}\makebox[14pt][c]{$\pm$}\makebox[34pt][c]{\num{0.0004}\rlap{*}} \\
\hline
\textbf{CAMEL-CLIP (Ours)} & \makebox[34pt][c]{\textbf{\num{0.8357}}}\makebox[14pt][c]{$\pm$}\makebox[34pt][c]{\textbf{\num{0.0012}}} & \makebox[34pt][c]{\textbf{\num{0.6748}}}\makebox[14pt][c]{$\pm$}\makebox[34pt][c]{\textbf{\num{0.0021}}} & \makebox[34pt][c]{\textbf{\num{0.7348}}}\makebox[14pt][c]{$\pm$}\makebox[34pt][c]{\textbf{\num{0.0024}}} \\
\hline
\end{tabular}}
\end{table}

% Response Table A2
\begin{table}[p]
\centering
\makebox[\textwidth][c]{%
\begin{minipage}{18.5cm}
\centering
\small
\setlength{\tabcolsep}{1.5pt}
\renewcommand{\arraystretch}{1.35}
\caption{\textbf{Downstream classification performance on CHB-MIT and MUMTAZ-2016.} This table reports performance on seizure detection on the CHB-MIT dataset and major depressive disorder classification on the MUMTAZ-2016 dataset under full fine-tuning and linear probing. Values are mean $\pm$ SD; - indicates unavailable linear probing results. $\dagger$ next to a model name indicates that its full fine-tuning results are reported from other papers. An asterisk (*) next to a performance value indicates a statistically significant difference from CAMEL-CLIP (two-sided Welch's t-test, n = 5 per model, Holm-adjusted p < 0.05); the direction of the difference is determined by the mean values. Holm correction is applied separately to eight comparison families defined by dataset (2) $\times$ evaluation setting (2) $\times$ metric (2), with nine comparisons per full fine-tuning family and five per linear probing family.}
\label{tab:cls_CHB}
\label{tab:cls_MUM}
\footnotesize
\begin{tabular}{@{}lcccccccc@{}}
\hline
\textbf{Method} & \multicolumn{4}{c}{\textbf{CHB-MIT dataset}} & \multicolumn{4}{c}{\textbf{MUMTAZ-2016 dataset}} \\
\cline{2-5}\cline{6-9}
 & \multicolumn{2}{c}{\textbf{Full fine-tuning}} & \multicolumn{2}{c}{\textbf{Linear probing}} & \multicolumn{2}{c}{\textbf{Full fine-tuning}} & \multicolumn{2}{c}{\textbf{Linear probing}} \\
\cline{2-3}\cline{4-5}\cline{6-7}\cline{8-9}
 & \makecell[c]{\textbf{Balanced}\\\textbf{accuracy}} & \textbf{AUC-PR} & \makecell[c]{\textbf{Balanced}\\\textbf{accuracy}} & \textbf{AUC-PR} & \makecell[c]{\textbf{Balanced}\\\textbf{accuracy}} & \textbf{AUC-PR} & \makecell[c]{\textbf{Balanced}\\\textbf{accuracy}} & \textbf{AUC-PR} \\
\hline
EEGNet \cite{lawhern2018eegnet}$\dagger$ & \makecell[c]{\tablenum[table-format=1.4]{0.5658}\makebox[0.45em][l]{}\\$\pm$ \tablenum[table-format=1.4]{0.0106}\makebox[0.45em][l]{*}} & \makecell[c]{\tablenum[table-format=1.4]{0.1914}\makebox[0.45em][l]{}\\$\pm$ \tablenum[table-format=1.4]{0.0182}\makebox[0.45em][l]{*}} & - & - & \makecell[c]{\tablenum[table-format=1.4]{0.9232}\makebox[0.45em][l]{}\\$\pm$ \tablenum[table-format=1.4]{0.0104}\makebox[0.45em][l]{}} & \makecell[c]{\tablenum[table-format=1.4]{0.9626}\makebox[0.45em][l]{}\\$\pm$ \tablenum[table-format=1.4]{0.0095}\makebox[0.45em][l]{}} & - & - \\
\hline
FFCL \cite{li2022motor}$\dagger$ & \makecell[c]{\tablenum[table-format=1.4]{0.6262}\makebox[0.45em][l]{}\\$\pm$ \tablenum[table-format=1.4]{0.0104}\makebox[0.45em][l]{*}} & \makecell[c]{\tablenum[table-format=1.4]{0.2049}\makebox[0.45em][l]{}\\$\pm$ \tablenum[table-format=1.4]{0.0346}\makebox[0.45em][l]{*}} & - & - & \makecell[c]{\tablenum[table-format=1.4]{0.9314}\makebox[0.45em][l]{}\\$\pm$ \tablenum[table-format=1.4]{0.0038}\makebox[0.45em][l]{*}} & \makecell[c]{\tablenum[table-format=1.4]{0.9717}\makebox[0.45em][l]{}\\$\pm$ \tablenum[table-format=1.4]{0.0021}\makebox[0.45em][l]{*}} & - & - \\
\hline
BIOT \cite{yang2023biot}$\dagger$ & \makecell[c]{\tablenum[table-format=1.4]{0.7068}\makebox[0.45em][l]{}\\$\pm$ \tablenum[table-format=1.4]{0.0457}\makebox[0.45em][l]{*}} & \makecell[c]{\tablenum[table-format=1.4]{0.3277}\makebox[0.45em][l]{}\\$\pm$ \tablenum[table-format=1.4]{0.0460}\makebox[0.45em][l]{}} & - & - & \makecell[c]{\tablenum[table-format=1.4]{0.9358}\makebox[0.45em][l]{}\\$\pm$ \tablenum[table-format=1.4]{0.0052}\makebox[0.45em][l]{*}} & \makecell[c]{\tablenum[table-format=1.4]{0.9736}\makebox[0.45em][l]{}\\$\pm$ \tablenum[table-format=1.4]{0.0034}\makebox[0.45em][l]{}} & - & - \\
\hline
LaBraM-Base \cite{jiang2024large}$\dagger$ & \makecell[c]{\tablenum[table-format=1.4]{0.7075}\makebox[0.45em][l]{}\\$\pm$ \tablenum[table-format=1.4]{0.0358}\makebox[0.45em][l]{*}} & \makecell[c]{\tablenum[table-format=1.4]{0.3287}\makebox[0.45em][l]{}\\$\pm$ \tablenum[table-format=1.4]{0.0402}\makebox[0.45em][l]{}} & - & - & \makecell[c]{\textbf{\tablenum[table-format=1.4]{0.9409}\makebox[0.45em][l]{}}\\\textbf{$\pm$ \tablenum[table-format=1.4]{0.0079}\makebox[0.45em][l]{*}}} & \makecell[c]{\tablenum[table-format=1.4]{0.9798}\makebox[0.45em][l]{}\\$\pm$ \tablenum[table-format=1.4]{0.0093}\makebox[0.45em][l]{}} & - & - \\
\hline
EEG-CLIP \cite{ndir2025eeg} & \makecell[c]{\tablenum[table-format=1.4]{0.6844}\makebox[0.45em][l]{}\\$\pm$ \tablenum[table-format=1.4]{0.0601}\makebox[0.45em][l]{*}} & \makecell[c]{\tablenum[table-format=1.4]{0.2191}\makebox[0.45em][l]{}\\$\pm$ \tablenum[table-format=1.4]{0.0819}\makebox[0.45em][l]{}} & \makecell[c]{\tablenum[table-format=1.4]{0.5000}\makebox[0.45em][l]{}\\$\pm$ \tablenum[table-format=1.4]{0.0000}\makebox[0.45em][l]{*}} & \makecell[c]{\tablenum[table-format=1.4]{0.0981}\makebox[0.45em][l]{}\\$\pm$ \tablenum[table-format=1.4]{0.0234}\makebox[0.45em][l]{*}} & \makecell[c]{\tablenum[table-format=1.4]{0.6583}\makebox[0.45em][l]{}\\$\pm$ \tablenum[table-format=1.4]{0.0096}\makebox[0.45em][l]{*}} & \makecell[c]{\tablenum[table-format=1.4]{0.6550}\makebox[0.45em][l]{}\\$\pm$ \tablenum[table-format=1.4]{0.0128}\makebox[0.45em][l]{*}} & \makecell[c]{\tablenum[table-format=1.4]{0.5031}\makebox[0.45em][l]{}\\$\pm$ \tablenum[table-format=1.4]{0.0047}\makebox[0.45em][l]{*}} & \makecell[c]{\tablenum[table-format=1.4]{0.5549}\makebox[0.45em][l]{}\\$\pm$ \tablenum[table-format=1.4]{0.1033}\makebox[0.45em][l]{*}} \\
\hline
CBraMod \cite{wang2024cbramod} & \makecell[c]{\tablenum[table-format=1.4]{0.6173}\makebox[0.45em][l]{}\\$\pm$ \tablenum[table-format=1.4]{0.0748}\makebox[0.45em][l]{*}} & \makecell[c]{\tablenum[table-format=1.4]{0.3779}\makebox[0.45em][l]{}\\$\pm$ \tablenum[table-format=1.4]{0.1412}\makebox[0.45em][l]{}} & \makecell[c]{\tablenum[table-format=1.4]{0.5744}\makebox[0.45em][l]{}\\$\pm$ \tablenum[table-format=1.4]{0.0157}\makebox[0.45em][l]{*}} & \makecell[c]{\textbf{\tablenum[table-format=1.4]{0.4169}\makebox[0.45em][l]{}}\\\textbf{$\pm$ \tablenum[table-format=1.4]{0.0201}\makebox[0.45em][l]{*}}} & \makecell[c]{\tablenum[table-format=1.4]{0.8874}\makebox[0.45em][l]{}\\$\pm$ \tablenum[table-format=1.4]{0.0206}\makebox[0.45em][l]{}} & \makecell[c]{\tablenum[table-format=1.4]{0.9763}\makebox[0.45em][l]{}\\$\pm$ \tablenum[table-format=1.4]{0.0060}\makebox[0.45em][l]{}} & \makecell[c]{\tablenum[table-format=1.4]{0.5109}\makebox[0.45em][l]{}\\$\pm$ \tablenum[table-format=1.4]{0.0020}\makebox[0.45em][l]{*}} & \makecell[c]{\tablenum[table-format=1.4]{0.9575}\makebox[0.45em][l]{}\\$\pm$ \tablenum[table-format=1.4]{0.0023}\makebox[0.45em][l]{*}} \\
\hline
MMM \cite{yi2023learning} & \makecell[c]{\tablenum[table-format=1.4]{0.7624}\makebox[0.45em][l]{}\\$\pm$ \tablenum[table-format=1.4]{0.0290}\makebox[0.45em][l]{*}} & \makecell[c]{\tablenum[table-format=1.4]{0.2570}\makebox[0.45em][l]{}\\$\pm$ \tablenum[table-format=1.4]{0.0585}\makebox[0.45em][l]{}} & \makecell[c]{\tablenum[table-format=1.4]{0.7750}\makebox[0.45em][l]{}\\$\pm$ \tablenum[table-format=1.4]{0.0444}\makebox[0.45em][l]{}} & \makecell[c]{\tablenum[table-format=1.4]{0.3016}\makebox[0.45em][l]{}\\$\pm$ \tablenum[table-format=1.4]{0.0751}\makebox[0.45em][l]{}} & \makecell[c]{\tablenum[table-format=1.4]{0.9018}\makebox[0.45em][l]{}\\$\pm$ \tablenum[table-format=1.4]{0.0337}\makebox[0.45em][l]{}} & \makecell[c]{\textbf{\tablenum[table-format=1.4]{0.9855}\makebox[0.45em][l]{}}\\\textbf{$\pm$ \tablenum[table-format=1.4]{0.0044}\makebox[0.45em][l]{}}} & \makecell[c]{\tablenum[table-format=1.4]{0.9248}\makebox[0.45em][l]{}\\$\pm$ \tablenum[table-format=1.4]{0.0153}\makebox[0.45em][l]{*}} & \makecell[c]{\tablenum[table-format=1.4]{0.9813}\makebox[0.45em][l]{}\\$\pm$ \tablenum[table-format=1.4]{0.0102}\makebox[0.45em][l]{}} \\
\hline
LUNA-Base \cite{doner2025luna} & \makecell[c]{\tablenum[table-format=1.4]{0.6267}\makebox[0.45em][l]{}\\$\pm$ \tablenum[table-format=1.4]{0.0646}\makebox[0.45em][l]{*}} & \makecell[c]{\tablenum[table-format=1.4]{0.3839}\makebox[0.45em][l]{}\\$\pm$ \tablenum[table-format=1.4]{0.0853}\makebox[0.45em][l]{}} & \makecell[c]{\tablenum[table-format=1.4]{0.5220}\makebox[0.45em][l]{}\\$\pm$ \tablenum[table-format=1.4]{0.0160}\makebox[0.45em][l]{*}} & \makecell[c]{\tablenum[table-format=1.4]{0.2384}\makebox[0.45em][l]{}\\$\pm$ \tablenum[table-format=1.4]{0.0395}\makebox[0.45em][l]{*}} & \makecell[c]{\tablenum[table-format=1.4]{0.9058}\makebox[0.45em][l]{}\\$\pm$ \tablenum[table-format=1.4]{0.0201}\makebox[0.45em][l]{}} & \makecell[c]{\tablenum[table-format=1.4]{0.9803}\makebox[0.45em][l]{}\\$\pm$ \tablenum[table-format=1.4]{0.0021}\makebox[0.45em][l]{}} & \makecell[c]{\tablenum[table-format=1.4]{0.8885}\makebox[0.45em][l]{}\\$\pm$ \tablenum[table-format=1.4]{0.0043}\makebox[0.45em][l]{*}} & \makecell[c]{\tablenum[table-format=1.4]{0.9568}\makebox[0.45em][l]{}\\$\pm$ \tablenum[table-format=1.4]{0.0030}\makebox[0.45em][l]{*}} \\
\hline
LUNA-Huge \cite{doner2025luna} & \makecell[c]{\tablenum[table-format=1.4]{0.5793}\makebox[0.45em][l]{}\\$\pm$ \tablenum[table-format=1.4]{0.0622}\makebox[0.45em][l]{*}} & \makecell[c]{\textbf{\tablenum[table-format=1.4]{0.4057}\makebox[0.45em][l]{}}\\\textbf{$\pm$ \tablenum[table-format=1.4]{0.0389}\makebox[0.45em][l]{}}} & \makecell[c]{\tablenum[table-format=1.4]{0.5366}\makebox[0.45em][l]{}\\$\pm$ \tablenum[table-format=1.4]{0.0264}\makebox[0.45em][l]{*}} & \makecell[c]{\tablenum[table-format=1.4]{0.3395}\makebox[0.45em][l]{}\\$\pm$ \tablenum[table-format=1.4]{0.0247}\makebox[0.45em][l]{}} & \makecell[c]{\tablenum[table-format=1.4]{0.8941}\makebox[0.45em][l]{}\\$\pm$ \tablenum[table-format=1.4]{0.0223}\makebox[0.45em][l]{}} & \makecell[c]{\tablenum[table-format=1.4]{0.9778}\makebox[0.45em][l]{}\\$\pm$ \tablenum[table-format=1.4]{0.0033}\makebox[0.45em][l]{}} & \makecell[c]{\tablenum[table-format=1.4]{0.9116}\makebox[0.45em][l]{}\\$\pm$ \tablenum[table-format=1.4]{0.0089}\makebox[0.45em][l]{*}} & \makecell[c]{\tablenum[table-format=1.4]{0.9750}\makebox[0.45em][l]{}\\$\pm$ \tablenum[table-format=1.4]{0.0064}\makebox[0.45em][l]{*}} \\
\hline
\makecell[l]{\textbf{CAMEL-CLIP}\\\textbf{(Ours)}} & \makecell[c]{\textbf{\tablenum[table-format=1.4]{0.8267}\makebox[0.45em][l]{}}\\\textbf{$\pm$ \tablenum[table-format=1.4]{0.0507}\makebox[0.45em][l]{}}} & \makecell[c]{\tablenum[table-format=1.4]{0.3813}\makebox[0.45em][l]{}\\$\pm$ \tablenum[table-format=1.4]{0.0649}\makebox[0.45em][l]{}} & \makecell[c]{\textbf{\tablenum[table-format=1.4]{0.7989}\makebox[0.45em][l]{}}\\\textbf{$\pm$ \tablenum[table-format=1.4]{0.0047}\makebox[0.45em][l]{}}} & \makecell[c]{\tablenum[table-format=1.4]{0.3690}\makebox[0.45em][l]{}\\$\pm$ \tablenum[table-format=1.4]{0.0048}\makebox[0.45em][l]{}} & \makecell[c]{\tablenum[table-format=1.4]{0.9129}\makebox[0.45em][l]{}\\$\pm$ \tablenum[table-format=1.4]{0.0084}\makebox[0.45em][l]{}} & \makecell[c]{\tablenum[table-format=1.4]{0.9813}\makebox[0.45em][l]{}\\$\pm$ \tablenum[table-format=1.4]{0.0044}\makebox[0.45em][l]{}} & \makecell[c]{\textbf{\tablenum[table-format=1.4]{0.9439}\makebox[0.45em][l]{}}\\\textbf{$\pm$ \tablenum[table-format=1.4]{0.0019}\makebox[0.45em][l]{}}} & \makecell[c]{\textbf{\tablenum[table-format=1.4]{0.9901}\makebox[0.45em][l]{}}\\\textbf{$\pm$ \tablenum[table-format=1.4]{0.0002}\makebox[0.45em][l]{}}} \\
\hline
\end{tabular}
\end{minipage}}
\end{table}

Finally, Table \ref{tab:cls_MUM} shows that CAMEL-CLIP achieves the highest mean balanced accuracy under linear probing on the MUMTAZ-2016 dataset. After performing Holm correction, its balanced accuracy is significantly higher than that of all five baselines. For CAMEL-CLIP, linear probing also yields higher mean balanced accuracy than full fine-tuning. These results support the transferability of the pretrained representations to this downstream task without updating the EEG encoder.

\FloatBarrier
\subsection{Ablation study and positional encoding comparison}
\label{sec5.4}

The proposed method consists of three main components: CAPE, DCP, and DCL. Because DCL requires the output of DCP, we group DCP and DCL as Dual-level Alignment (DLA). In Experiment 3, we conduct an ablation study covering all four combinations of including or excluding CAPE and DLA. CAPE provides information that differentiates channels, while DLA uses this information for channel-level and sample-level EEG--text representation learning. The TUAB results are reported in Table \ref{abl_TUAB}; the CHB-MIT and MUMTAZ-2016 results are reported together in Table \ref{abl_CHB}.

As shown in Table \ref{abl_TUAB}, applying DLA substantially improves mean performance on the dataset used for pretraining. CAMEL-CLIP, which includes all three modules, shows differences across tasks but achieves lower overall mean performance than the configuration with DLA alone. CAPE alone yields lower mean performance than the configuration with all modules removed. However, the TUAB evaluation uses the same dataset and channel configuration as pretraining and therefore does not involve the channel configuration changes encountered during downstream transfer. In this setting, the model can learn representations for the channels consistently available during pretraining, which may limit the additional benefit of the explicit channel identity information provided by CAPE. CAPE is intended to preserve channel identity when the available channels or montage change across datasets.

The downstream dataset results in Table \ref{abl_CHB} highlight the importance of CAPE and its complementary relationship with DLA. On the CHB-MIT dataset, DLA alone does not improve performance, and CAPE alone reduces AUC-PR. However, using CAPE and DLA together achieves the highest AUC-PR, showing the benefit of combining channel identity information with EEG--text alignment. On the MUMTAZ-2016 dataset, either CAPE or DLA alone improves performance over the configuration with all modules removed, but using them together further improves both balanced accuracy and AUC-PR. These results suggest that the proposed modules play complementary roles and that combining them can improve downstream performance.

% Response Table A10
\begin{table}[!t]
\centering
\small
\setlength{\tabcolsep}{8pt}
\renewcommand{\arraystretch}{1.35}
\caption{\textbf{Ablation study on the TUAB dataset.} This table summarizes text-based classification performance on the TUAB dataset, where CAPE, DCP, and DCL denote channel attribute-based positional encoding, dynamic channel projection, and dual-level contrastive learning, respectively. $\checkmark$ indicates inclusion and $\times$ indicates removal; DLA (Dual-level Alignment) denotes the combined configuration of DCP and DCL; both modules are included or removed together. When CAPE is removed, the asymmetric conditional positional encoding (ACPE) originally used in CBraMod is used instead. The final row represents CAMEL-CLIP.}
\label{abl_TUAB}
\footnotesize
\makebox[\textwidth][c]{%
\begin{tabular}{@{}lccccc@{}}
\hline
\textbf{CAPE} & \textbf{DLA} & \textbf{Pathological} & \textbf{Gender} & \textbf{Age} & \textbf{Average} \\
\hline
$\times$ & $\times$ & \tablenum[table-format=1.4]{0.7270} & \textbf{\tablenum[table-format=1.4]{0.6633}} & \tablenum[table-format=1.4]{0.4874} & \tablenum[table-format=1.4]{0.6259} \\
\hline
$\checkmark$ & $\times$ & \tablenum[table-format=1.4]{0.7172} & \tablenum[table-format=1.4]{0.6545} & \tablenum[table-format=1.4]{0.4407} & \tablenum[table-format=1.4]{0.6041} \\
\hline
$\times$ & $\checkmark$ & \tablenum[table-format=1.4]{0.8097} & \tablenum[table-format=1.4]{0.6540} & \textbf{\tablenum[table-format=1.4]{0.7372}} & \textbf{\tablenum[table-format=1.4]{0.7336}} \\
\hline
$\checkmark$ & $\checkmark$ & \textbf{\tablenum[table-format=1.4]{0.8158}} & \tablenum[table-format=1.4]{0.6463} & \tablenum[table-format=1.4]{0.7125} & \tablenum[table-format=1.4]{0.7249} \\
\hline
\end{tabular}}
\end{table}

% Response Table A11
\begin{table}[htbp]
\centering
\makebox[\textwidth][c]{%
\begin{minipage}{18.5cm}
\centering
\small
\setlength{\tabcolsep}{5pt}
\renewcommand{\arraystretch}{1.4}
\caption{\textbf{Ablation study on the CHB-MIT and MUMTAZ-2016 datasets.} This table summarizes classification performance on the CHB-MIT and MUMTAZ-2016 datasets, where CAPE, DCP, and DCL denote channel attribute-based positional encoding, dynamic channel projection, and dual-level contrastive learning, respectively. All models are evaluated under the linear probing setting. Values are mean $\pm$ SD. $\checkmark$ indicates inclusion and $\times$ indicates removal; DLA (Dual-level Alignment) denotes the combined configuration of DCP and DCL; both modules are included or removed together. When CAPE is removed, the asymmetric conditional positional encoding (ACPE) originally used in CBraMod is used instead. The final row represents CAMEL-CLIP. An asterisk (*) indicates a statistically significant difference from CAMEL-CLIP (two-sided Welch's t-test, n = 5 per model, Holm-adjusted p < 0.05). Holm correction is applied separately to each dataset and metric across three CAMEL-CLIP--ablation comparisons; the direction is determined by the mean values.}
\label{abl_CHB}
\label{abl_MUM}
\begin{tabular}{@{}cccccc@{}}
\hline
\textbf{CAPE} & \textbf{DLA} & \multicolumn{2}{c}{\textbf{CHB-MIT dataset}} & \multicolumn{2}{c}{\textbf{MUMTAZ-2016 dataset}} \\
\cline{3-4}\cline{5-6}
 & & \textbf{Balanced accuracy} & \textbf{AUC-PR} & \textbf{Balanced accuracy} & \textbf{AUC-PR} \\
\hline
$\times$ & $\times$ & \makebox[34pt][c]{\num{0.7889}}\makebox[14pt][c]{$\pm$}\makebox[34pt][c]{\num{0.0075}} & \makebox[34pt][c]{\num{0.2971}}\makebox[14pt][c]{$\pm$}\makebox[34pt][c]{\num{0.0060}\rlap{*}} & \makebox[34pt][c]{\num{0.8937}}\makebox[14pt][c]{$\pm$}\makebox[34pt][c]{\num{0.0007}\rlap{*}} & \makebox[34pt][c]{\num{0.9610}}\makebox[14pt][c]{$\pm$}\makebox[34pt][c]{\num{0.0003}\rlap{*}} \\
\hline
$\checkmark$ & $\times$ & \makebox[34pt][c]{\textbf{\num{0.8002}}}\makebox[14pt][c]{$\pm$}\makebox[34pt][c]{\textbf{\num{0.0071}}} & \makebox[34pt][c]{\num{0.2425}}\makebox[14pt][c]{$\pm$}\makebox[34pt][c]{\num{0.0071}\rlap{*}} & \makebox[34pt][c]{\num{0.9153}}\makebox[14pt][c]{$\pm$}\makebox[34pt][c]{\num{0.0016}\rlap{*}} & \makebox[34pt][c]{\num{0.9807}}\makebox[14pt][c]{$\pm$}\makebox[34pt][c]{\num{0.0005}\rlap{*}} \\
\hline
$\times$ & $\checkmark$ & \makebox[34pt][c]{\num{0.7067}}\makebox[14pt][c]{$\pm$}\makebox[34pt][c]{\num{0.0048}\rlap{*}} & \makebox[34pt][c]{\num{0.1981}}\makebox[14pt][c]{$\pm$}\makebox[34pt][c]{\num{0.0160}\rlap{*}} & \makebox[34pt][c]{\num{0.9086}}\makebox[14pt][c]{$\pm$}\makebox[34pt][c]{\num{0.0016}\rlap{*}} & \makebox[34pt][c]{\num{0.9768}}\makebox[14pt][c]{$\pm$}\makebox[34pt][c]{\num{0.0007}\rlap{*}} \\
\hline
$\checkmark$ & $\checkmark$ & \makebox[34pt][c]{\num{0.7989}}\makebox[14pt][c]{$\pm$}\makebox[34pt][c]{\num{0.0047}} & \makebox[34pt][c]{\textbf{\num{0.3690}}}\makebox[14pt][c]{$\pm$}\makebox[34pt][c]{\textbf{\num{0.0048}}} & \makebox[34pt][c]{\textbf{\num{0.9439}}}\makebox[14pt][c]{$\pm$}\makebox[34pt][c]{\textbf{\num{0.0019}}} & \makebox[34pt][c]{\textbf{\num{0.9901}}}\makebox[14pt][c]{$\pm$}\makebox[34pt][c]{\textbf{\num{0.0002}}} \\
\hline
\end{tabular}
\end{minipage}}
\end{table}

% Response Table A12

To further evaluate the contribution of CAPE, we compare it with Asymmetric Conditional Positional Encoding (ACPE) from CBraMod \cite{wang2024cbramod} and 3D coordinate Positional Encoding (PE) adapted from CoordConformer \cite{patil2024coordconformer}. ACPE dynamically generates positional information from the spatiotemporal characteristics of EEG features, whereas 3D coordinate PE uses electrode coordinates to represent spatial relationships across different channel configurations. We replace only the channel positional encoding in CAMEL-CLIP, keeping all other components and training conditions unchanged. Each model is trained independently, and downstream evaluation is performed with a frozen pretrained EEG encoder and a trained classifier. As shown in Table \ref{tab:pe_comparison}, CAPE achieves the highest mean balanced accuracy and AUC-PR on both CHB-MIT and MUMTAZ-2016. These results suggest that CAPE supports more effective transfer to downstream tasks than the baseline positional encoding methods.

% Response Table A1
\begin{table}[htbp]
\centering
\makebox[\textwidth][c]{%
\begin{minipage}{185mm}
\centering
\small
\setlength{\tabcolsep}{2pt}
\renewcommand{\arraystretch}{1.35}
\caption{\textbf{Comparison of positional encoding variants on the CHB-MIT and MUMTAZ-2016 datasets under linear probing with a frozen encoder.} Values are mean $\pm$ SD. Bold performance values indicate the highest mean in each column.}
\label{tab:pe_comparison}
\begin{tabular*}{185mm}{@{\extracolsep{\fill}}lcccc@{}}
\hline
\textbf{Method} & \multicolumn{2}{c}{\textbf{CHB-MIT dataset}} & \multicolumn{2}{c}{\textbf{MUMTAZ-2016 dataset}} \\
\cline{2-3}\cline{4-5}
 & \textbf{Balanced accuracy} & \textbf{AUC-PR} & \textbf{Balanced accuracy} & \textbf{AUC-PR} \\
\hline
with ACPE \cite{wang2024cbramod} & \tablenum[table-format=1.4]{0.7067} $\pm$ \tablenum[table-format=1.4]{0.0048} & \tablenum[table-format=1.4]{0.1981} $\pm$ \tablenum[table-format=1.4]{0.0160} & \tablenum[table-format=1.4]{0.9086} $\pm$ \tablenum[table-format=1.4]{0.0016} & \tablenum[table-format=1.4]{0.9768} $\pm$ \tablenum[table-format=1.4]{0.0007} \\
\hline
with 3D coordinate PE \cite{patil2024coordconformer} & \tablenum[table-format=1.4]{0.7938} $\pm$ \tablenum[table-format=1.4]{0.0103} & \tablenum[table-format=1.4]{0.3094} $\pm$ \tablenum[table-format=1.4]{0.0204} & \tablenum[table-format=1.4]{0.9345} $\pm$ \tablenum[table-format=1.4]{0.0035} & \tablenum[table-format=1.4]{0.9853} $\pm$ \tablenum[table-format=1.4]{0.0002} \\

\hline
with CAPE (Ours) & \textbf{\tablenum[table-format=1.4]{0.7989} $\pm$ \tablenum[table-format=1.4]{0.0047}} & \textbf{\tablenum[table-format=1.4]{0.3690} $\pm$ \tablenum[table-format=1.4]{0.0048}} & \textbf{\tablenum[table-format=1.4]{0.9439} $\pm$ \tablenum[table-format=1.4]{0.0019}} & \textbf{\tablenum[table-format=1.4]{0.9901} $\pm$ \tablenum[table-format=1.4]{0.0002}} \\
\hline
\end{tabular*}
\end{minipage}}
\end{table}

\FloatBarrier
\begin{table}[H]
\centering
\small
\setlength{\tabcolsep}{18pt}
\renewcommand{\arraystretch}{1.35}
\caption{\textbf{Five held-out channel configurations.} Each set excludes one electrode from each of the frontal, central, temporal, parietal, and occipital regions during pretraining.}
\label{tab:heldout}
\begin{tabular*}{0.72\linewidth}{@{\extracolsep{\fill}}lc@{}}
\hline
\textbf{Set} & \textbf{Excluded channels} \\
\hline
A & F7, C3, T6, P4, O2 \\
B & FZ, C3, T5, PZ, O2 \\
C & F7, C4, T4, P3, O1 \\
D & F4, CZ, T6, P3, O1 \\
E & F8, C4, T3, P3, O1 \\
\hline
\end{tabular*}
\end{table}

\subsection{Channel representation analysis and robustness to channel heterogeneity}
\label{sec5.2}

Experiment 4 analyzes channel representations and evaluates robustness to channel heterogeneity. We examine the consistency of embeddings from the same channel and the separation between different channels through visualizations and quantitative analyses. We also evaluate classification performance under asymmetric channel mismatch. To assess whether the observed trends depend on the choice of excluded channels, we construct five held-out channel configurations, as listed in Table \ref{tab:heldout} \cite{ma2026codebrain,sirca2026beyond}. Each configuration excludes one channel from each of the frontal, central, temporal, parietal, and occipital regions during pretraining. Using multiple configurations allows us to examine whether the observed trends are consistent across different channel selections rather than relying on a single configuration.

First, UMAP visualizations \cite{mcinnes2018umap} confirm that the CCL objective forces the encoder to extract highly individualized channel embeddings. Figure \ref{fig6} reveals that the CCL-trained model forms tightly separated clusters, unlike the non-CCL baseline, proving its ability to capture channel-specific semantics.

\clearpage
\begingroup
\setlength{\intextsep}{6pt}
\begin{figure}[H]
    \centering
    \begin{minipage}[b]{0.48\linewidth}
    \centering
    \subfloat[Train without CCL\label{fig6a}]{
        \includegraphics[width=\linewidth,height=0.22\textheight,keepaspectratio]{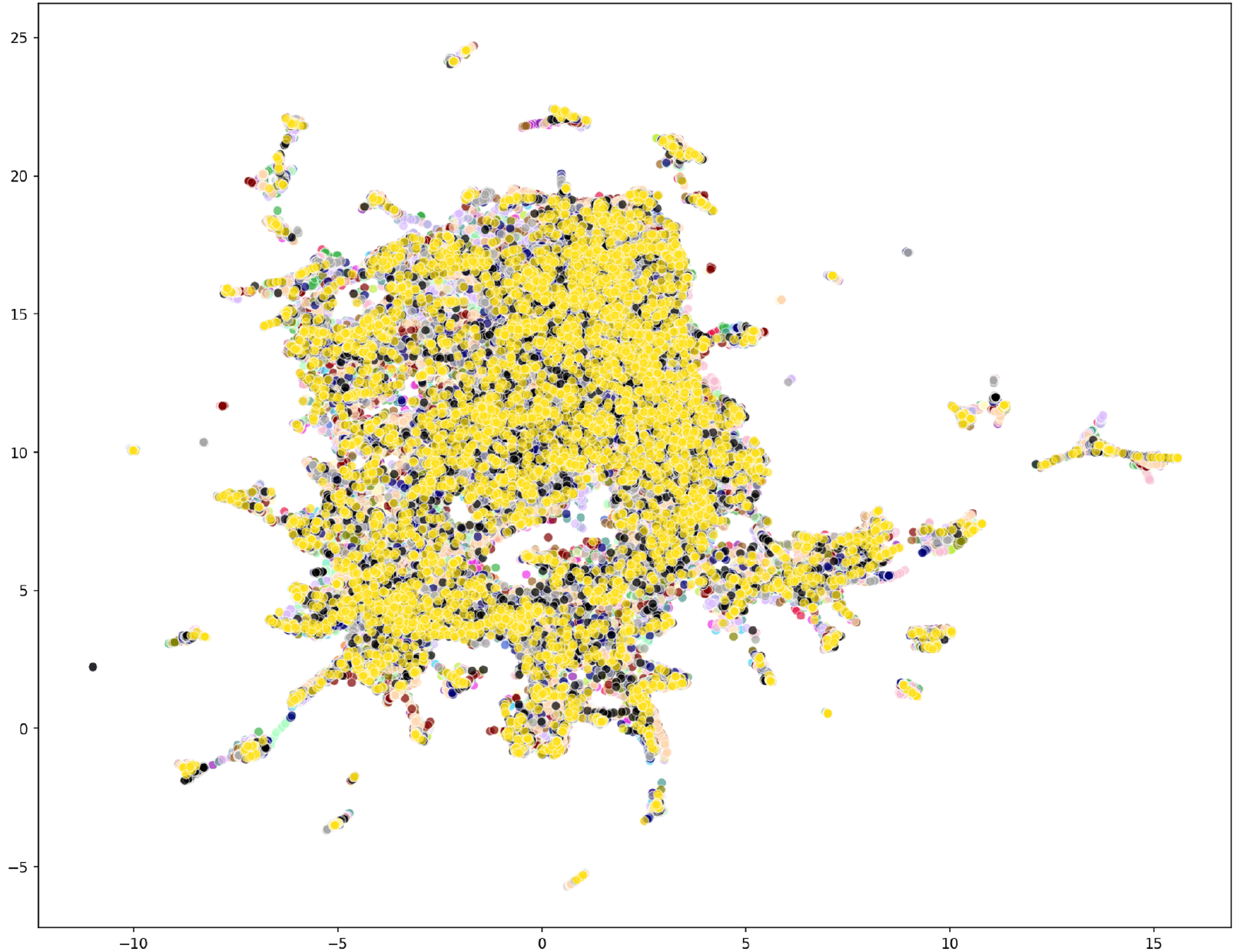}
    }
    \end{minipage}
    \hfill
    \begin{minipage}[b]{0.48\linewidth}
    \centering
    \subfloat[Train with CCL\label{fig6b}]{
        \includegraphics[width=\linewidth,height=0.22\textheight,keepaspectratio]{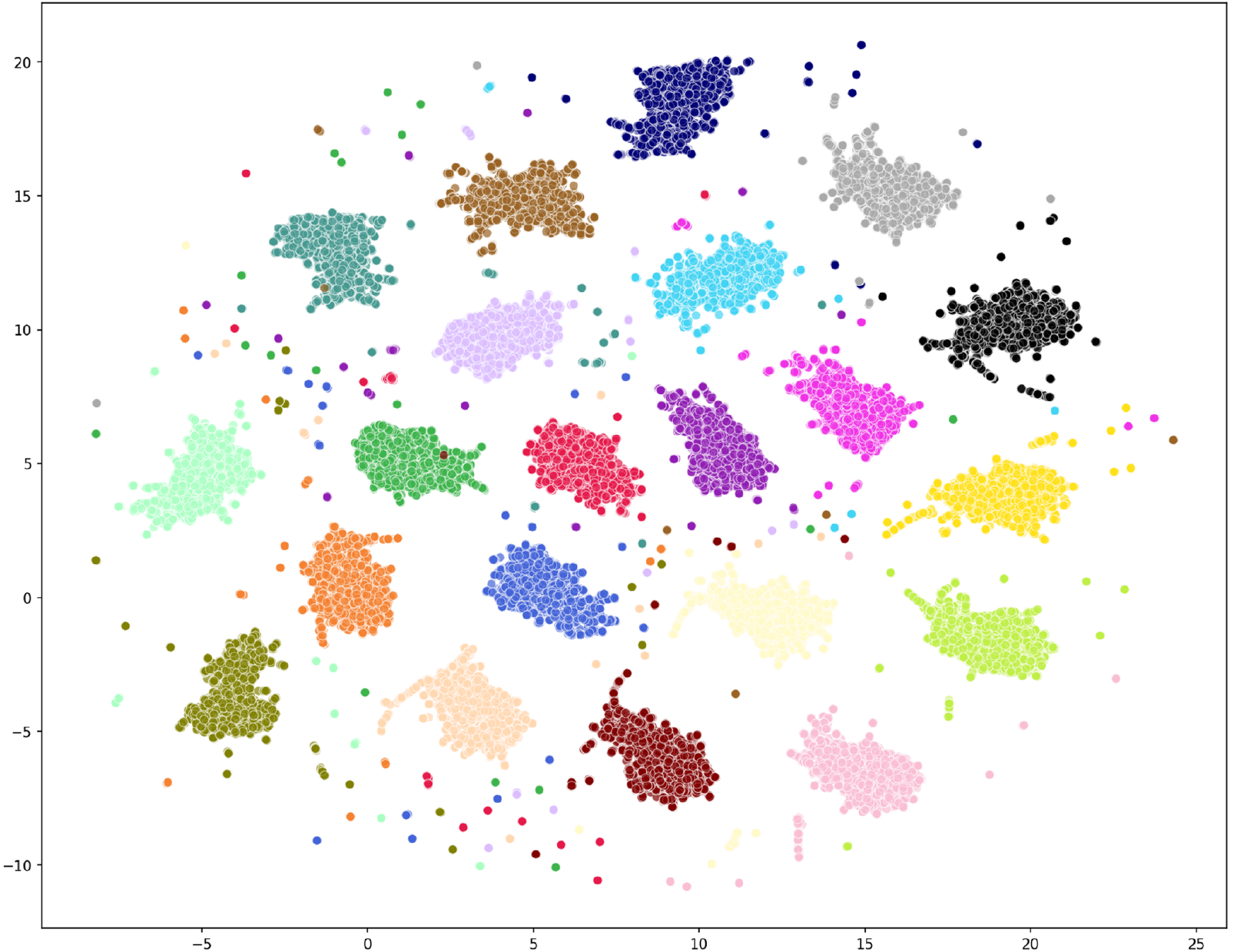}
    }
    \end{minipage}

    \vspace{0.3em}
    \begin{minipage}[b]{0.48\linewidth}
    \centering
    \subfloat[Train with all channels\label{fig6c}]{
        \includegraphics[width=\linewidth,height=0.22\textheight,keepaspectratio]{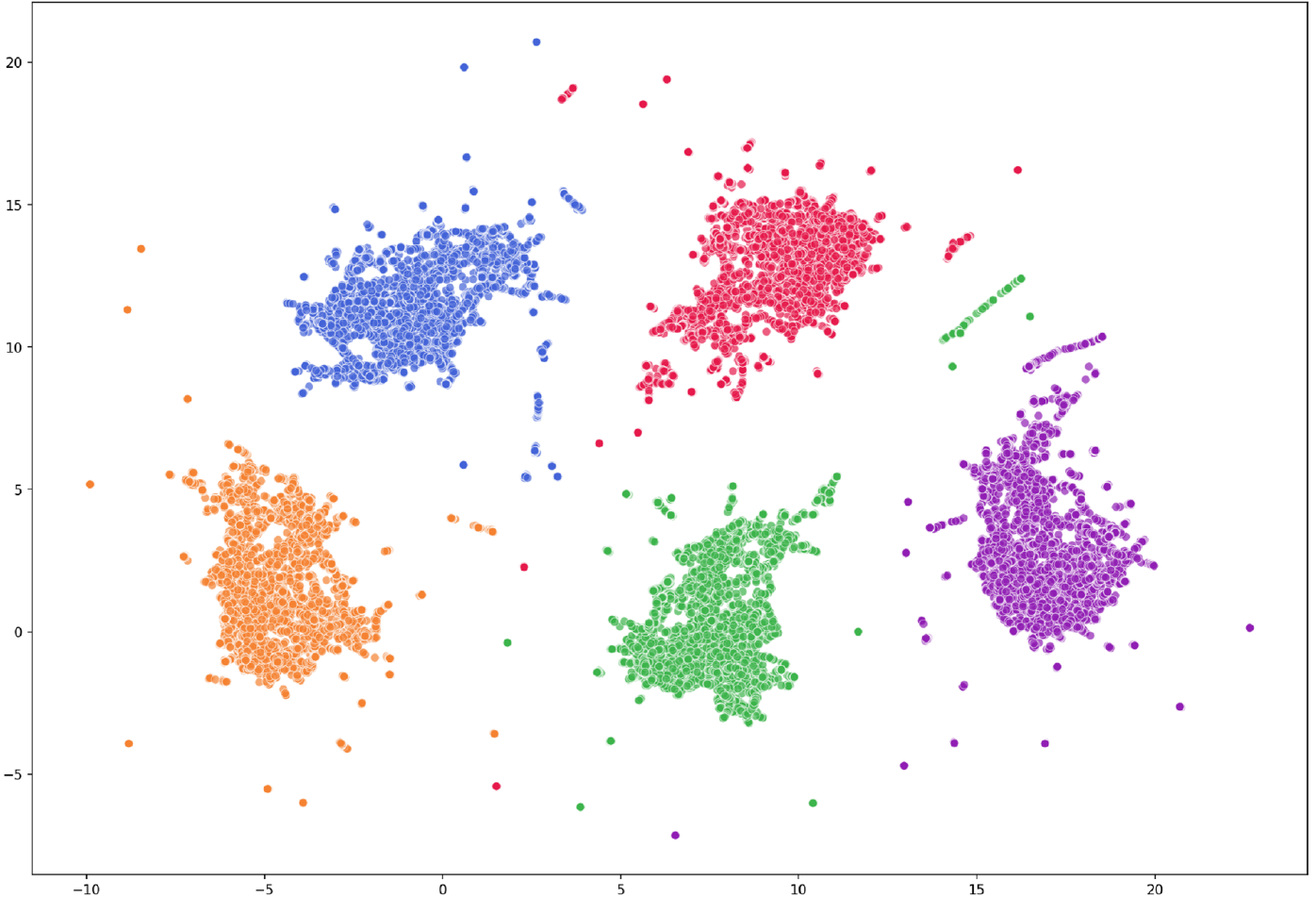}
    }
    \end{minipage}
    \hfill
    \begin{minipage}[b]{0.48\linewidth}
    \centering
    \subfloat[Train without five channels\label{fig6d}]{
        \includegraphics[width=\linewidth,height=0.22\textheight,keepaspectratio]{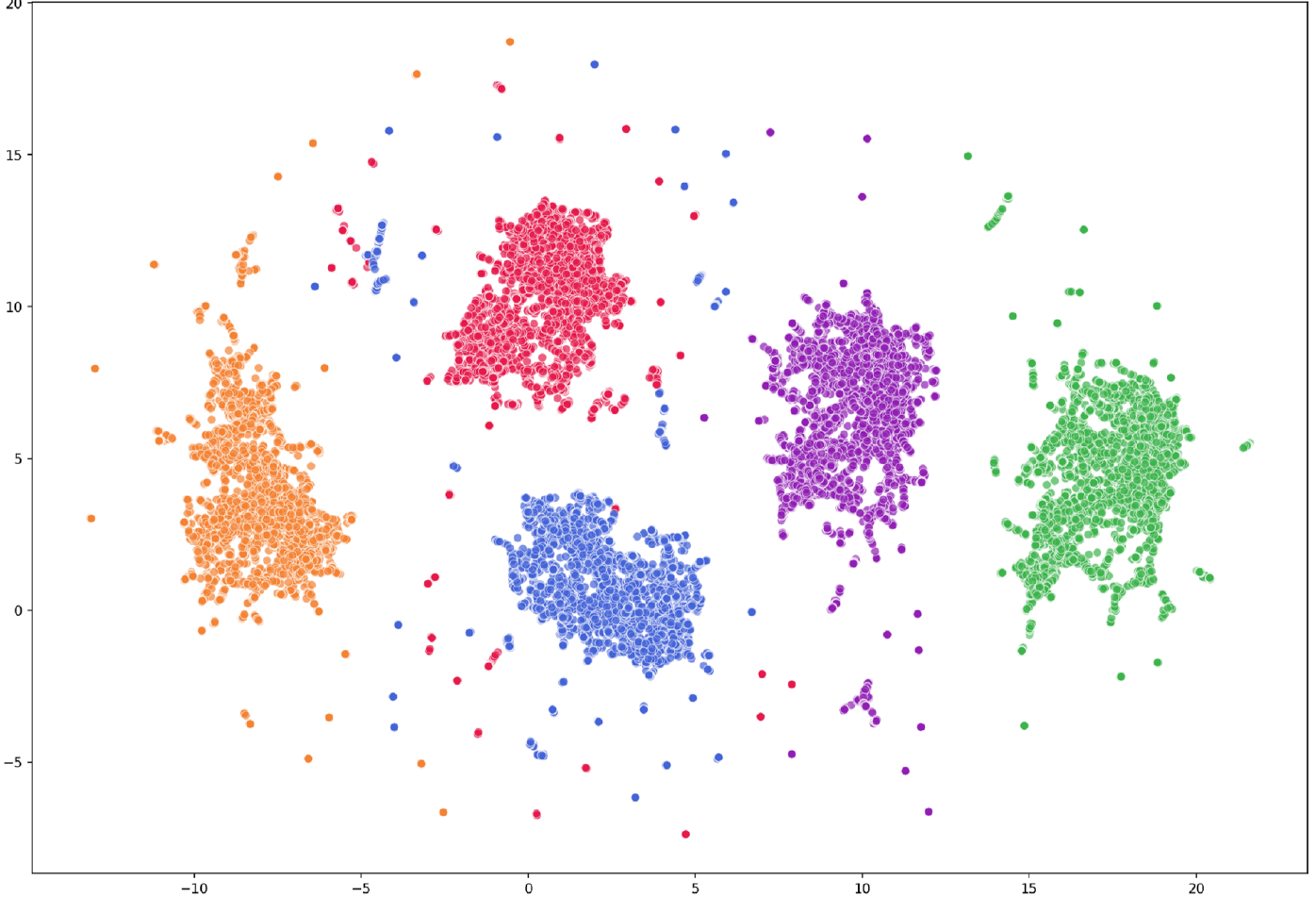}
    }
    \end{minipage}
    \caption{\textbf{Channel-wise EEG embedding visualizations on the TUAB dataset.} (a) Channel-wise EEG embeddings from the model trained without channel-level contrastive learning (CCL). (b) Channel-wise EEG embeddings from the model trained with CCL. (c) Channel-wise EEG embeddings for five channels (\textit{i.e., }F7, C3, T6, P4, and O2) from the model trained using all channels. (d) Channel-wise EEG embeddings from the model trained excluding the five channels.}

    \label{fig6}
\end{figure}
\endgroup

Next, we examine zero-shot generalization using held-out Set A from Table \ref{tab:heldout}, which excludes F7, C3, T6, P4, and O2 during pretraining on TUAB. Comparing the embedding distributions of a model pretrained using all channels (Figure \ref{fig6}(c)) with a model pretrained without these five channels (Figure \ref{fig6}(d)) reveals nearly identical clustering patterns. This indicates that our model intrinsically learns distinctive profiles for entirely unseen channels.

\begin{figure}[H]
    \centering
    \begin{minipage}[b]{0.48\linewidth}
    \centering
    \subfloat[Train without CCL\label{fig7a}]{
        \includegraphics[width=\linewidth,height=0.25\textheight,keepaspectratio]{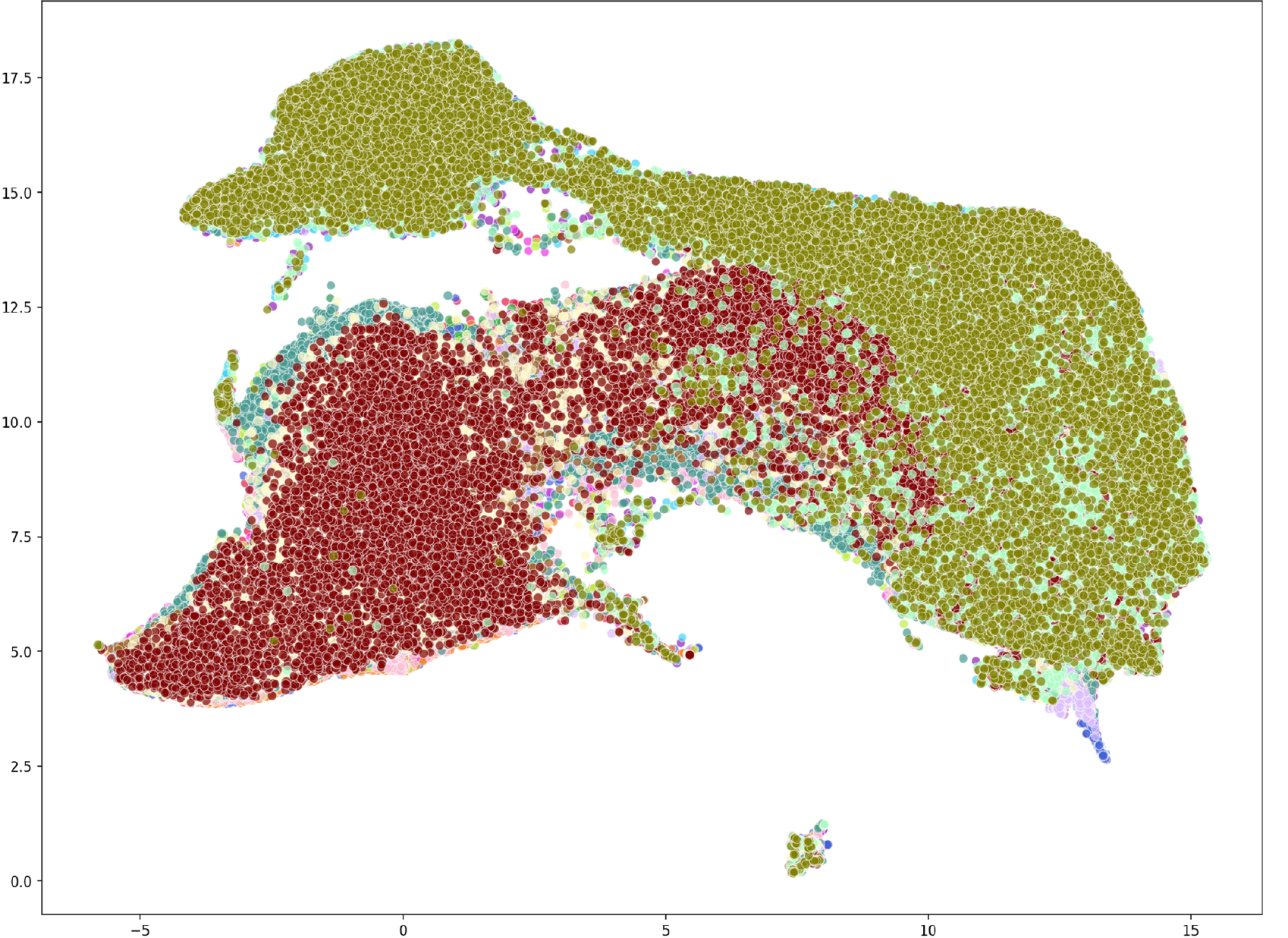}
    }
    \end{minipage}
    \hfill
    \begin{minipage}[b]{0.48\linewidth}
    \centering
    \subfloat[Train with CCL\label{fig7b}]{
        \includegraphics[width=\linewidth,height=0.25\textheight,keepaspectratio]{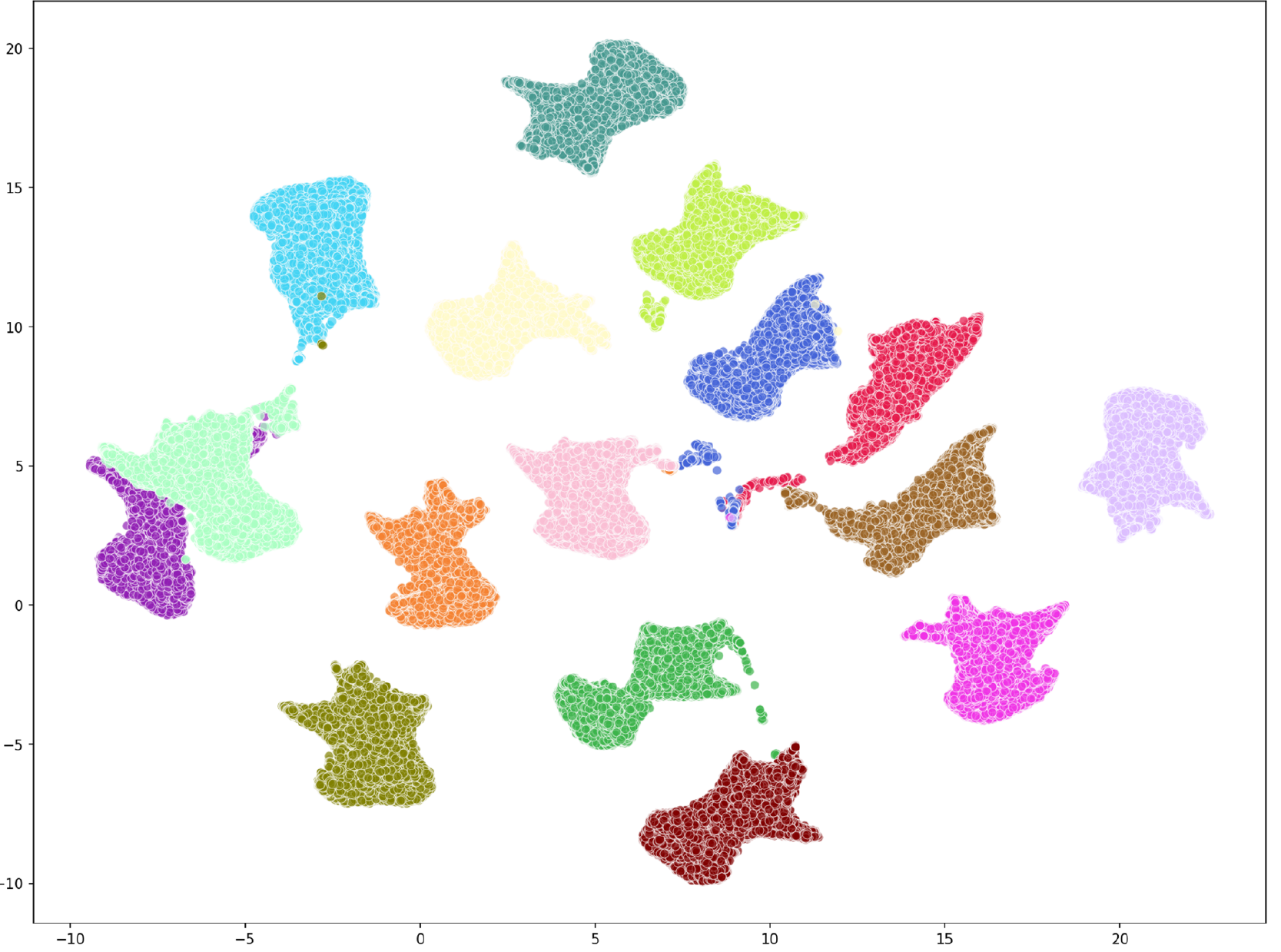}
    }
    \end{minipage}
    \caption{\textbf{Channel-wise EEG embedding visualizations on the CHB-MIT dataset.} (a) Channel-wise EEG embeddings from the model trained without channel-level contrastive learning (CCL). (b) Channel-wise EEG embeddings from the model trained with CCL.}
    \label{fig7}
\end{figure}

We further validate CCL's adaptability on the CHB-MIT dataset, which employs a different montage. Consistent with previous results, Figure \ref{fig7} shows that CCL is crucial for achieving clear channel separation, demonstrating its broad applicability across datasets with significant channel heterogeneity.

\clearpage

To complement the UMAP visualizations, Figure \ref{fig8} compares cross-window cosine similarity matrices for models trained without and with CCL under held-out Set A. This analysis examines whether embeddings from the same channel remain similar to one another and dissimilar to those from other channels across different windows. The five channels labeled in red are unseen channels excluded from pretraining, whereas those labeled in black are seen channels included in pretraining. Compared with the model trained without CCL, the model trained with CCL shows more clearly that embeddings from the same channel have higher cosine similarity across different EEG windows than embeddings from different channels. This pattern is also observed for unseen channels. Each panel is independently min-max normalized; colors therefore indicate relative structure within a panel and cannot be compared across panels in absolute cosine similarity units. Figure \ref{fig9} visualizes cosine similarities between the channel prototypes of the unseen channels C3 and P4 and those of other channels, arranged according to electrode positions. Each prototype is obtained by averaging the L2-normalized embeddings of a channel across all 5,796 TUAB evaluation windows and then L2-normalizing the mean. The displayed cosine similarities are not min--max normalized. Figure \ref{fig9}(a) shows that the C3 prototype exhibits high similarity with those of C4 and FP2, while Figure \ref{fig9}(b) shows that the P4 prototype exhibits high similarity with those of P3 and F4.

\clearpage
\begingroup
\setlength{\intextsep}{6pt}
\begin{figure}[H]
    \centering
    \begin{minipage}[b]{0.49\linewidth}
    \centering
    \subfloat[Train without CCL\label{fig8a}]{
        \includegraphics[width=\linewidth,height=0.24\textheight,keepaspectratio]{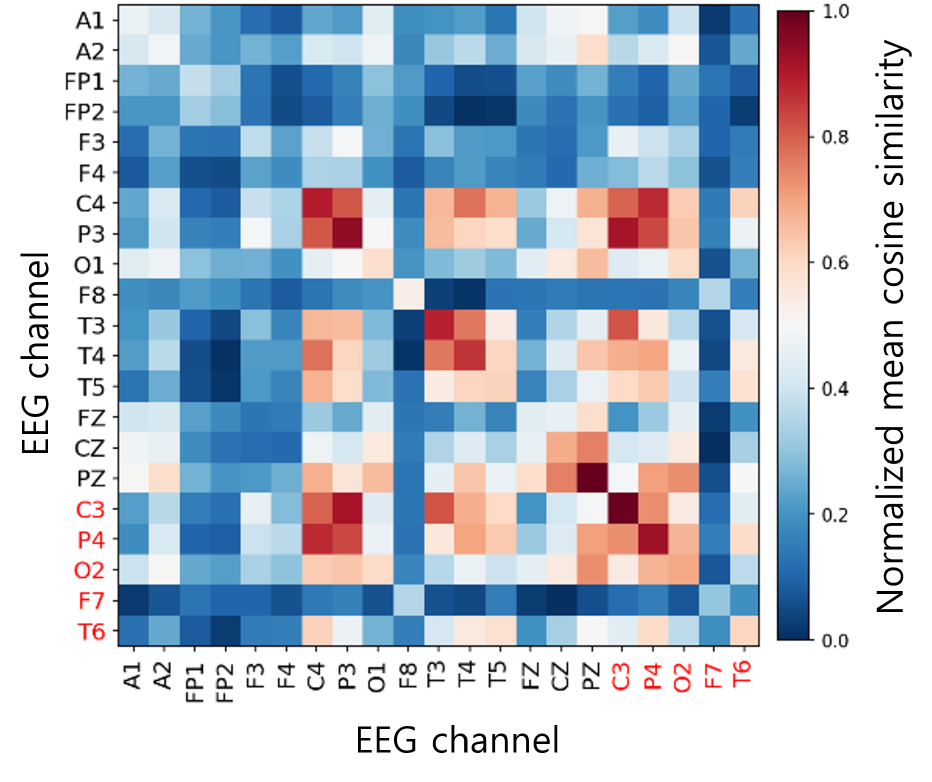}
    }
    \end{minipage}
    \hfill
    \begin{minipage}[b]{0.49\linewidth}
    \centering
    \subfloat[Train with CCL\label{fig8b}]{
        \includegraphics[width=\linewidth,height=0.24\textheight,keepaspectratio]{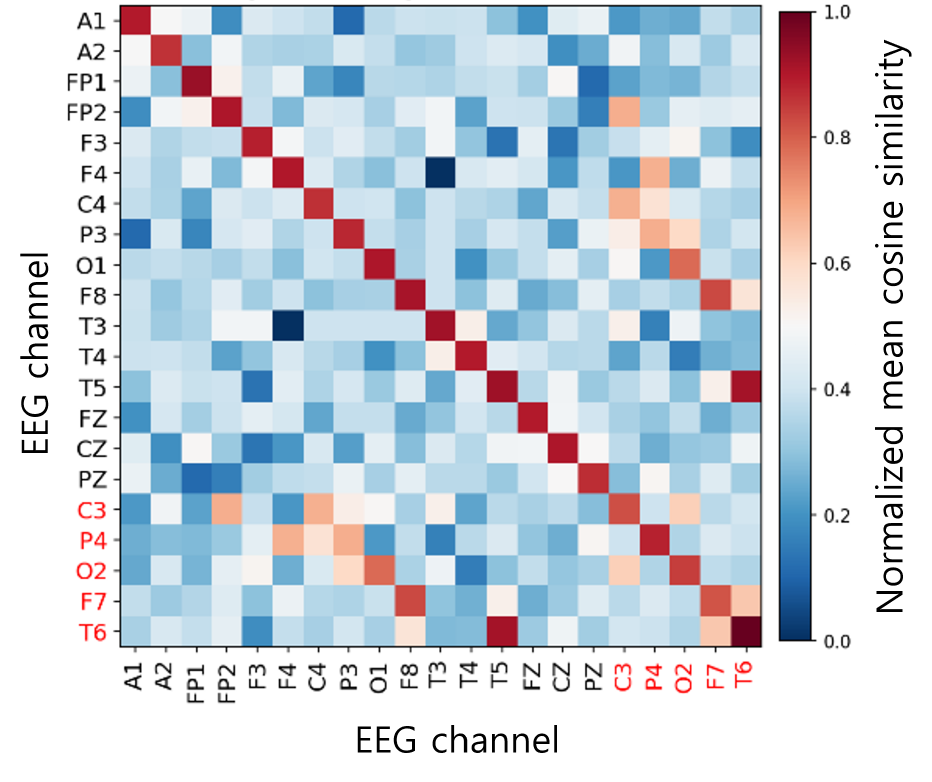}
    }
    \end{minipage}
    \caption{\textbf{Cross-window cosine similarity matrices between channel-level EEG embeddings from models trained (a) without CCL and (b) with CCL under held-out channel Set A.} Red channel labels identify channels excluded during training (i.e., unseen channels). Black channel labels identify channels included during training (i.e., seen channels). Each panel is independently min-max normalized to visualize patterns; colors cannot be compared across panels in absolute cosine similarity units.}
    \label{fig8}
\end{figure}

\begin{figure}[H]
    \centering
    \begin{minipage}[b]{0.495\linewidth}
    \centering
    \subfloat[Channel similarity to C3\label{fig9a}]{
        \includegraphics[width=\linewidth,height=0.24\textheight,keepaspectratio]{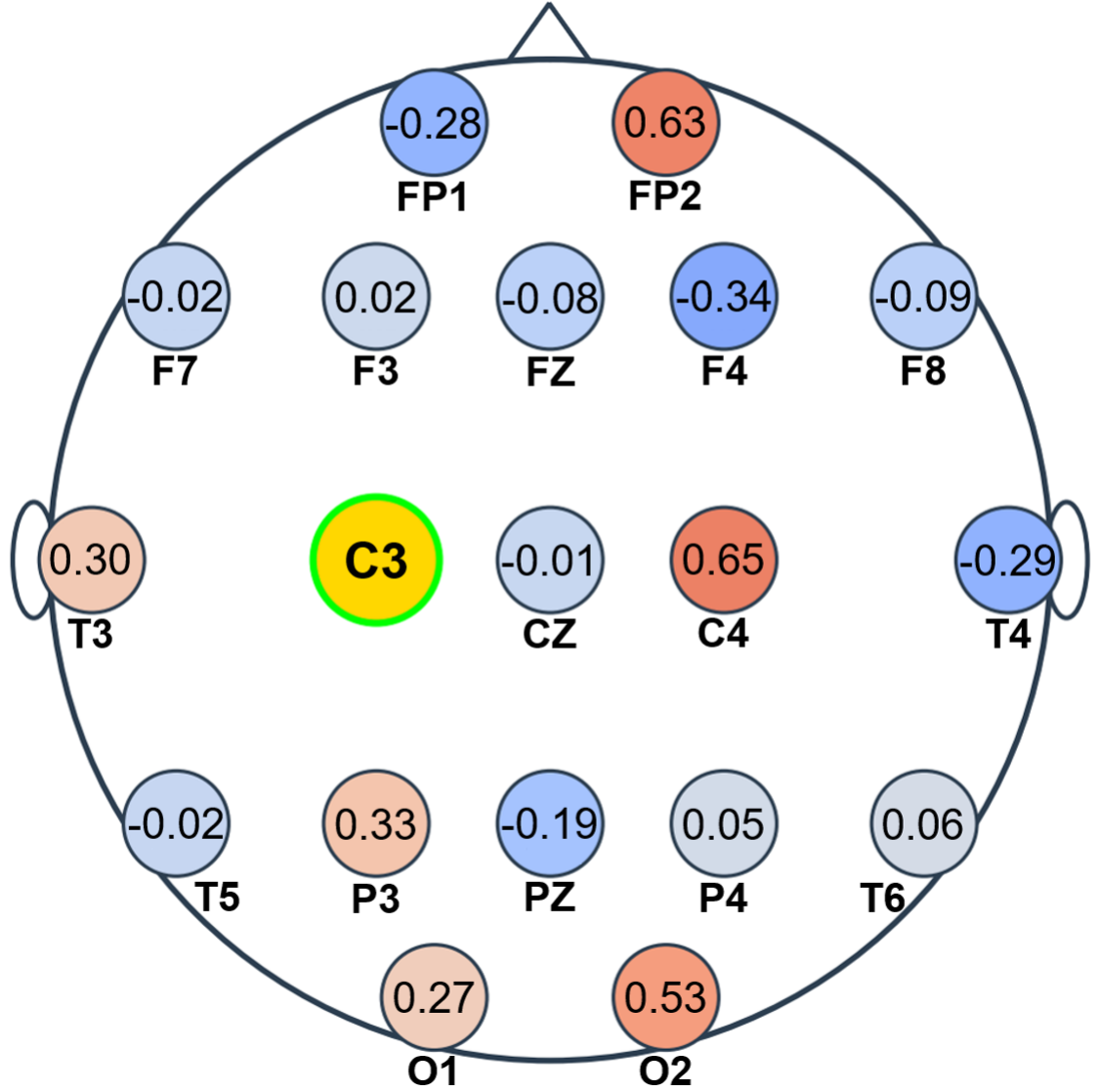}
    }
    \end{minipage}
    \hfill
    \begin{minipage}[b]{0.495\linewidth}
    \centering
    \subfloat[Channel similarity to P4\label{fig9b}]{
        \includegraphics[width=\linewidth,height=0.24\textheight,keepaspectratio]{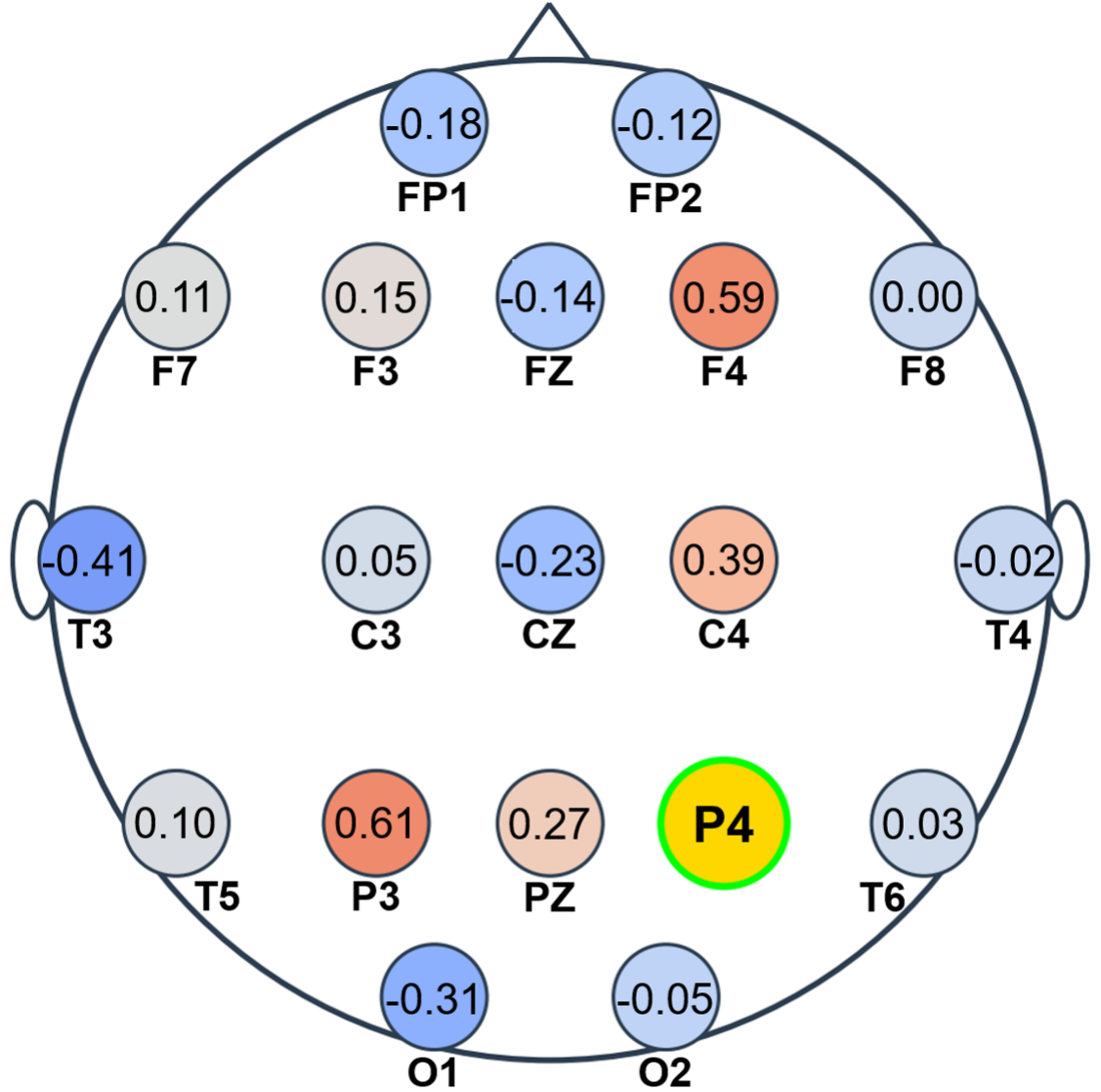}
    }
    \end{minipage}
    \caption{\textbf{Visualization of cosine similarity between channel prototypes.} Channel prototypes are computed from the TUAB evaluation windows using the model pretrained while excluding five channels (F7, C3, T6, P4, and O2; held-out Set A). (a) Cosine similarity between the C3 prototype and those of other channels. (b) Cosine similarity between the P4 prototype and those of other channels. Values are displayed at the corresponding electrode positions without min--max normalization.}
    \label{fig9}
\end{figure}
\endgroup

\clearpage

\begin{table}[H]
\centering
\small
\setlength{\tabcolsep}{5pt}
\renewcommand{\arraystretch}{1.25}
\caption{\textbf{Quantitative analysis of EEG channel embedding vectors from models trained without and with CCL.} Intra-/inter-channel cosine distance, Vendi Score, and Silhouette Score are reported for seen (16), unseen (5), and all (21) channels. Values are mean $\pm$ sample SD across five held-out channel configurations, with the pretraining seed fixed. Lower intra-channel distance and higher values of the other metrics are preferred.}
\label{tab:channel_metrics}
\makebox[\textwidth][c]{%
\begin{tabular}{@{}>{\raggedright\arraybackslash}p{3.6cm}lcc@{}}
\hline
\textbf{Metric} & \textbf{Channel group} & \textbf{without CCL} & \textbf{with CCL} \\
\hline
\multirow{3}{3.6cm}{\raggedright Intra-channel cosine distance} & Seen (16) & \tablenum[table-format=1.4]{0.7084} $\pm$ \tablenum[table-format=1.4]{0.4034} & \tablenum[table-format=2.4]{0.5751} $\pm$ \tablenum[table-format=2.4]{0.0591} \\
 & Unseen (5) & \tablenum[table-format=1.4]{0.7123} $\pm$ \tablenum[table-format=1.4]{0.4032} & \tablenum[table-format=2.4]{0.5985} $\pm$ \tablenum[table-format=2.4]{0.0627} \\
 & All (21) & \tablenum[table-format=1.4]{0.7093} $\pm$ \tablenum[table-format=1.4]{0.4033} & \tablenum[table-format=2.4]{0.5807} $\pm$ \tablenum[table-format=2.4]{0.0563} \\
\hline
\multirow{3}{3.6cm}{\raggedright Inter-channel cosine distance} & Seen (16) & \tablenum[table-format=1.4]{0.7389} $\pm$ \tablenum[table-format=1.4]{0.4149} & \tablenum[table-format=2.4]{1.0091} $\pm$ \tablenum[table-format=2.4]{0.0056} \\
 & Unseen (5) & \tablenum[table-format=1.4]{0.7374} $\pm$ \tablenum[table-format=1.4]{0.4140} & \tablenum[table-format=2.4]{0.9913} $\pm$ \tablenum[table-format=2.4]{0.0301} \\
 & All (21) & \tablenum[table-format=1.4]{0.7380} $\pm$ \tablenum[table-format=1.4]{0.4144} & \tablenum[table-format=2.4]{0.9988} $\pm$ \tablenum[table-format=2.4]{0.0056} \\
\hline
\multirow{3}{3.6cm}{\raggedright Vendi Score} & Seen (16) & \tablenum[table-format=1.4]{2.1987} $\pm$ \tablenum[table-format=1.4]{0.7712} & \tablenum[table-format=2.4]{10.8827} $\pm$ \tablenum[table-format=2.4]{0.3404} \\
 & Unseen (5) & \tablenum[table-format=1.4]{1.7881} $\pm$ \tablenum[table-format=1.4]{0.4759} & \tablenum[table-format=2.4]{4.1178} $\pm$ \tablenum[table-format=2.4]{0.4033} \\
 & All (21) & \tablenum[table-format=1.4]{2.2113} $\pm$ \tablenum[table-format=1.4]{0.7525} & \tablenum[table-format=2.4]{10.5991} $\pm$ \tablenum[table-format=2.4]{0.3507} \\
\hline
\multirow{3}{3.6cm}{\raggedright Silhouette Score} & Seen (16) & \tablenum[table-format=1.4]{0.0136} $\pm$ \tablenum[table-format=1.4]{0.1491} & \tablenum[table-format=2.4]{0.2649} $\pm$ \tablenum[table-format=2.4]{0.0583} \\
 & Unseen (5) & \tablenum[table-format=1.4]{0.0127} $\pm$ \tablenum[table-format=1.4]{0.1638} & \tablenum[table-format=2.4]{0.1299} $\pm$ \tablenum[table-format=2.4]{0.0641} \\
 & All (21) & \tablenum[table-format=1.4]{0.0134} $\pm$ \tablenum[table-format=1.4]{0.1525} & \tablenum[table-format=2.4]{0.2327} $\pm$ \tablenum[table-format=2.4]{0.0528} \\
\hline
\end{tabular}}
\end{table}

To quantify these properties, we evaluate intra-/inter-channel cosine distance, Vendi Score, and Silhouette Score across the five held-out channel configurations in Table \ref{tab:heldout}. Intra-/inter-channel distances assess the consistency of embeddings within a channel and the separation between channels. Vendi Score \cite{friedman2023vendi} evaluates the diversity of channel embeddings across different samples, while Silhouette Score \cite{rousseeuw1987silhouettes} evaluates whether embeddings from the same channel are closer to one another than to those from other channels. As shown in Table \ref{tab:channel_metrics}, the model trained with CCL exhibits lower mean intra-channel distance and higher mean inter-channel distance than the model trained without CCL for seen channels, unseen channels, and all channels combined. Together with the higher Silhouette Score, these results indicate that, in the model trained with CCL, embeddings from the same channel are more closely grouped across EEG windows and better separated from those of other channels. Vendi Score was also higher for the model trained with CCL, which suggests reduced redundancy and greater representational diversity among channel embeddings.

\clearpage
\begin{table}[H]
\centering
\small
\setlength{\tabcolsep}{1.5pt}
\renewcommand{\arraystretch}{1.55}
\setlength{\extrarowheight}{3pt}
\caption{\textbf{Asymmetric channel-mismatch evaluation.} Source and target balanced accuracy are reported separately for each direction of channel removal. Source denotes the channel configuration used to train the classifier, and Target denotes the configuration used for evaluation without retraining. The arrow indicates transfer from Source to Target. ``Left'' and ``Right'' denote configurations with left-hemisphere channels (F3, C3, T3, P3, and O1) and right-hemisphere channels (F4, C4, T4, P4, and O2) excluded, respectively. Thus, Left $\rightarrow$ Right denotes training with left-hemisphere channels excluded and evaluation with right-hemisphere channels excluded; Right $\rightarrow$ Left denotes the reverse direction. Values for each transfer direction are mean $\pm$ SD. Average Source balanced accuracy and Average Target balanced accuracy are reported as mean $\pm$ sample SD across the ten results pooled from both transfer directions (five seeds per direction). These SDs reflect both seed variability and differences between transfer directions.}
\label{tab:asymmetric}
\fontsize{9}{11}\selectfont
\begin{tabular*}{\textwidth}{@{\extracolsep{\fill}}lcccccc@{}}
\hline
\textbf{Model} & \multicolumn{2}{c}{\textbf{Left $\boldsymbol{\rightarrow}$ Right}} & \multicolumn{2}{c}{\textbf{Right $\boldsymbol{\rightarrow}$ Left}} & \multicolumn{2}{c}{\textbf{Average}} \\[3pt]
\cline{2-3}\cline{4-5}\cline{6-7}
 & \textbf{Source} & \textbf{Target} & \textbf{Source} & \textbf{Target} & \textbf{Source} & \textbf{Target} \\[3pt]
\hline
CBraMod \cite{wang2024cbramod} & \makecell[c]{\tablenum[table-format=1.4]{0.5104}\\$\pm$ \tablenum[table-format=1.4]{0.0070}} & \makecell[c]{\tablenum[table-format=1.4]{0.5066}\\$\pm$ \tablenum[table-format=1.4]{0.0064}} & \makecell[c]{\tablenum[table-format=1.4]{0.5188}\\$\pm$ \tablenum[table-format=1.4]{0.0135}} & \makecell[c]{\tablenum[table-format=1.4]{0.5194}\\$\pm$ \tablenum[table-format=1.4]{0.0131}} & \makecell[c]{\tablenum[table-format=1.4]{0.5146}\\$\pm$ \tablenum[table-format=1.4]{0.0111}} & \makecell[c]{\tablenum[table-format=1.4]{0.5130}\\$\pm$ \tablenum[table-format=1.4]{0.0118}} \\[3pt]
\hline
MMM \cite{yi2023learning} & \makecell[c]{\tablenum[table-format=1.4]{0.9097}\\$\pm$ \tablenum[table-format=1.4]{0.0195}} & \makecell[c]{\tablenum[table-format=1.4]{0.8566}\\$\pm$ \tablenum[table-format=1.4]{0.0525}} & \makecell[c]{\textbf{\tablenum[table-format=1.4]{0.9451}}\\\textbf{$\pm$ \tablenum[table-format=1.4]{0.0182}}} & \makecell[c]{\tablenum[table-format=1.4]{0.8654}\\$\pm$ \tablenum[table-format=1.4]{0.0448}} & \makecell[c]{\tablenum[table-format=1.4]{0.9274}\\$\pm$ \tablenum[table-format=1.4]{0.0258}} & \makecell[c]{\tablenum[table-format=1.4]{0.8610}\\$\pm$ \tablenum[table-format=1.4]{0.0462}} \\[3pt]
\hline
LUNA-Base \cite{doner2025luna} & \makecell[c]{\tablenum[table-format=1.4]{0.8898}\\$\pm$ \tablenum[table-format=1.4]{0.0086}} & \makecell[c]{\tablenum[table-format=1.4]{0.8250}\\$\pm$ \tablenum[table-format=1.4]{0.0305}} & \makecell[c]{\tablenum[table-format=1.4]{0.8715}\\$\pm$ \tablenum[table-format=1.4]{0.0039}} & \makecell[c]{\tablenum[table-format=1.4]{0.8816}\\$\pm$ \tablenum[table-format=1.4]{0.0075}} & \makecell[c]{\tablenum[table-format=1.4]{0.8807}\\$\pm$ \tablenum[table-format=1.4]{0.0115}} & \makecell[c]{\tablenum[table-format=1.4]{0.8533}\\$\pm$ \tablenum[table-format=1.4]{0.0364}} \\[3pt]

\hline
\makecell[l]{\textbf{CAMEL-CLIP}\\\textbf{(Ours)}} & \makecell[c]{\textbf{\tablenum[table-format=1.4]{0.9381}}\\\textbf{$\pm$ \tablenum[table-format=1.4]{0.0007}}} & \makecell[c]{\textbf{\tablenum[table-format=1.4]{0.9300}}\\\textbf{$\pm$ \tablenum[table-format=1.4]{0.0014}}} & \makecell[c]{\tablenum[table-format=1.4]{0.9389}\\$\pm$ \tablenum[table-format=1.4]{0.0028}} & \makecell[c]{\textbf{\tablenum[table-format=1.4]{0.9249}}\\\textbf{$\pm$ \tablenum[table-format=1.4]{0.0041}}} & \makecell[c]{\textbf{\tablenum[table-format=1.4]{0.9385}}\\\textbf{$\pm$ \tablenum[table-format=1.4]{0.0020}}} & \makecell[c]{\textbf{\tablenum[table-format=1.4]{0.9275}}\\\textbf{$\pm$ \tablenum[table-format=1.4]{0.0039}}} \\[3pt]
\hline
\end{tabular*}
\end{table}

Finally, we evaluate robustness to asymmetric channel mismatch on the \mbox{MUMTAZ-2016} dataset. We exclude either the left-hemisphere channels F3, C3, T3, P3, and O1 or their right-hemisphere counterparts F4, C4, T4, P4, and O2, removing both the signals and channel names from the input. Both configurations retain 14 channels and share nine channels. We keep the pretrained encoder frozen and train a classifier under one configuration, then apply it to the opposite configuration without retraining.

Table \ref{tab:asymmetric} reports source and target balanced accuracy for both transfer directions, comparing CAMEL-CLIP with CBraMod, MMM, and LUNA-Base under the same protocol. Source denotes the channel configuration used to train the classifier, whereas Target denotes the opposite configuration used for evaluation without classifier retraining. CAMEL-CLIP achieves the highest target balanced accuracy in both directions, with a maximum decrease of 1.40 percentage points relative to Source. These results show that CAMEL-CLIP maintains both high target balanced accuracy and a small performance decrease under the evaluated asymmetric channel mismatch conditions.

Overall, the visualizations and quantitative analyses indicate that CCL promotes consistent representations of the same channel across EEG samples and clearer separation between different channels, including channels excluded from pretraining. The asymmetric channel-mismatch evaluation further shows that CAMEL-CLIP maintains high classification performance when the channel configuration changes without classifier retraining. These results support the effectiveness of the proposed framework in learning channel-specific representations and handling channel heterogeneity.

\FloatBarrier

\FloatBarrier

\subsection{EEG-text alignment evaluation}
\label{sec:alignment}

Experiment 5 evaluates EEG--text alignment quality and the contribution of the text modality to model performance through three complementary experiments: comparison with EEG-only pretraining, cosine similarity analysis of positive and negative EEG--text pairs, and EEG--text retrieval.

First, we compare CAMEL-CLIP with an EEG-only model under linear probing. In the EEG-only model, we replace the EEG--text contrastive learning used for multimodal alignment of channel-level and sample-level representations with EEG--EEG contrastive learning. Specifically, the global and channel-level text embeddings are replaced with EEG-based global and channel-level reference embeddings, respectively, while all other components remain unchanged. The global reference embedding is computed by averaging EEG global embeddings associated with the same text report. The channel-level reference embedding is computed by averaging embeddings of the same channel across the entire training set, regardless of the associated report. The EEG embeddings used to generate these two reference embeddings were updated each epoch along with the EEG encoder. Table \ref{tab:eeg_only} shows that CAMEL-CLIP achieves higher mean balanced accuracy and AUC-PR than the EEG-only model under all evaluation settings. On the CHB-MIT dataset, for example, AUC-PR increases from 0.2072 to 0.3690. These findings suggest the contribution of the text modality to the performance improvements achieved by CAMEL-CLIP.

\clearpage
\begin{table}[H]
\centering
\small
\setlength{\tabcolsep}{5pt}
\renewcommand{\arraystretch}{1.35}
\caption{\textbf{Comparison of EEG-only and EEG-text pretraining (CAMEL-CLIP, Ours) under linear probing on the MUMTAZ-2016 and CHB-MIT datasets and the three TUAB tasks.} Values are mean $\pm$ SD.}
\label{tab:eeg_only}
\makebox[\textwidth][c]{%
\begin{tabular}{@{}>{\raggedright\arraybackslash}p{3.1cm}lcc@{}}
\hline
\textbf{Task} & \textbf{Model} & \textbf{Balanced accuracy} & \textbf{AUC-PR} \\
\hline
\multirow{2}{3.1cm}{MUMTAZ-2016 dataset} & EEG-only & \tablenum[table-format=1.4]{0.9216} $\pm$ \tablenum[table-format=1.4]{0.0033} & \tablenum[table-format=1.4]{0.9848} $\pm$ \tablenum[table-format=1.4]{0.0004} \\
\cline{2-4}
 & CAMEL-CLIP (Ours) & \textbf{\tablenum[table-format=1.4]{0.9439} $\pm$ \tablenum[table-format=1.4]{0.0019}} & \textbf{\tablenum[table-format=1.4]{0.9901} $\pm$ \tablenum[table-format=1.4]{0.0002}} \\
\hline
\multirow{2}{3.1cm}{CHB-MIT dataset} & EEG-only & \tablenum[table-format=1.4]{0.7958} $\pm$ \tablenum[table-format=1.4]{0.0185} & \tablenum[table-format=1.4]{0.2072} $\pm$ \tablenum[table-format=1.4]{0.0473} \\
\cline{2-4}
 & CAMEL-CLIP (Ours) & \textbf{\tablenum[table-format=1.4]{0.7989} $\pm$ \tablenum[table-format=1.4]{0.0047}} & \textbf{\tablenum[table-format=1.4]{0.3690} $\pm$ \tablenum[table-format=1.4]{0.0048}} \\
\hline
\multirow{2}{3.1cm}{TUAB dataset : pathological} & EEG-only & \tablenum[table-format=1.4]{0.7936} $\pm$ \tablenum[table-format=1.4]{0.0016} & \tablenum[table-format=1.4]{0.8772} $\pm$ \tablenum[table-format=1.4]{0.0014} \\
\cline{2-4}
 & CAMEL-CLIP (Ours) & \textbf{\tablenum[table-format=1.4]{0.8357} $\pm$ \tablenum[table-format=1.4]{0.0012}} & \textbf{\tablenum[table-format=1.4]{0.9106} $\pm$ \tablenum[table-format=1.4]{0.0022}} \\
\hline
\multirow{2}{3.1cm}{TUAB dataset : gender} & EEG-only & \tablenum[table-format=1.4]{0.6152} $\pm$ \tablenum[table-format=1.4]{0.0080} & \tablenum[table-format=1.4]{0.6899} $\pm$ \tablenum[table-format=1.4]{0.0106} \\
\cline{2-4}
 & CAMEL-CLIP (Ours) & \textbf{\tablenum[table-format=1.4]{0.6748} $\pm$ \tablenum[table-format=1.4]{0.0021}} & \textbf{\tablenum[table-format=1.4]{0.7470} $\pm$ \tablenum[table-format=1.4]{0.0016}} \\
\hline
\multirow{2}{3.1cm}{TUAB dataset : age} & EEG-only & \tablenum[table-format=1.4]{0.7127} $\pm$ \tablenum[table-format=1.4]{0.0077} & \tablenum[table-format=1.4]{0.7755} $\pm$ \tablenum[table-format=1.4]{0.0031} \\
\cline{2-4}
 & CAMEL-CLIP (Ours) & \textbf{\tablenum[table-format=1.4]{0.7348} $\pm$ \tablenum[table-format=1.4]{0.0024}} & \textbf{\tablenum[table-format=1.4]{0.8167} $\pm$ \tablenum[table-format=1.4]{0.0020}} \\
\hline
\end{tabular}}
\end{table}

In the second experiment, we compare the cosine similarity of positive and negative EEG--text pairs in the embedding space. Positive pairs are defined as an EEG sample and its corresponding text report from the same recording. Negative pairs are defined as an EEG sample and a text report from a different recording, excluding reports with identical text to the corresponding report. We use the sample-level text representation as the text vector on the TUAB test set. Because each text report corresponds to multiple EEG samples, we construct a representative EEG vector by averaging the sample-level EEG representations associated with that report. In Table \ref{tab:pair_similarity}, the mean cosine similarity is 0.6256 for positive pairs and 0.3439 for negative pairs, a difference of 0.2817. These results indicate that CAMEL-CLIP assigns higher similarity to positive EEG--text pairs than to negative pairs, suggesting cross-modal alignment between EEG and text representations.

% Response Table A3

% Response Table A4
\clearpage
\begin{table}[H]
\centering
\small
\setlength{\tabcolsep}{12pt}
\renewcommand{\arraystretch}{1.35}
\caption{\textbf{Cosine similarity between positive and negative EEG--text pairs in the original 1,408-dimensional embedding space.} Positive pairs share the same recording; negative pairs include all combinations with different text report texts, excluding identical text report strings. Values are mean $\pm$ SD across pairs, rather than across training runs or confidence intervals.}
\label{tab:pair_similarity}
\begin{tabular}{@{}lc@{}}
\hline
\textbf{Pair type} & \textbf{Cosine similarity (mean $\boldsymbol{\pm}$ SD)} \\
\hline
Positive pairs & 0.6256 $\pm$ 0.1412 \\
Negative pairs & 0.3439 $\pm$ 0.2373 \\
\hline
\end{tabular}
\end{table}

\label{sec5.5}
Finally, we perform an EEG--text retrieval experiment to further assess EEG--text alignment from another perspective. This task retrieves relevant text reports using EEG as a query, providing an indirect measure of alignment between EEG and text representations. We evaluate retrieval on the three TUAB tasks---pathological status, gender, and age group---using EEG-CLIP as the baseline. Relevance is defined by whether the query and the retrieved text report match on the attribute being evaluated. We report Precision@K, Recall@K, nDCG@K, and mAP, with K = 1, 5, 10, and 100 for the top-K metrics. Precision@K measures the proportion of relevant text reports among the top K results. Recall@K measures the proportion of relevant text reports in the full ranking that are retrieved within the top K results \cite{thakur2021beir}. mAP assesses retrieval quality across the full ranking \cite{khattab2020colbert}, while nDCG@K gives greater weight to relevant text reports ranked higher within the top K results \cite{santhanam2022colbertv2}. All metrics are calculated for each query window and then averaged.

As shown in Table \ref{ret}, CAMEL-CLIP achieves higher Precision@K and nDCG@K than EEG-CLIP at every K across all three tasks. Recall@K also generally improves. In particular, mAP increases for gender and age group. These results suggest improved retrieval of text reports matching these attributes across the full ranking, as well as among the top results. CAMEL-CLIP achieves Precision@1 values that are 3.08, 10.09, and 9.39 percentage points higher than those of EEG-CLIP on the pathological, gender, and age tasks, respectively.

\clearpage
\begin{table}[H]
\centering
\small
\setlength{\tabcolsep}{5pt}
\renewcommand{\arraystretch}{1.35}
\caption{\textbf{EEG--text retrieval performance on the TUAB dataset.} Precision@K, Recall@K, nDCG@K, and mean average precision (mAP) are reported for the pathological, gender, and age tasks. Relevance is defined by agreement between the query EEG label and the retrieved text report label for the corresponding task. Metrics are computed for each query window and then averaged. K is set to 1, 5, 10, and 100 for the top-K metrics; mAP is computed over the full retrieval ranking.}
\label{ret}
\footnotesize
\makebox[\textwidth][c]{%
\begin{tabular}{@{}llccccc@{}}
\hline
\textbf{Task} & \textbf{Method} & $\boldsymbol{K}$ & \textbf{Precision@$\boldsymbol{K}$} & \textbf{Recall@$\boldsymbol{K}$} & \textbf{nDCG@$\boldsymbol{K}$} & \textbf{mAP} \\
\hline
\multirow{8}{*}{Pathological} & \multirow{4}{*}{EEG-CLIP \cite{ndir2025eeg}} & 1 & \tablenum[table-format=1.4]{0.7874} & \textbf{\tablenum[table-format=1.4]{0.0008}} & \tablenum[table-format=1.4]{0.7874} & \multirow{4}{*}{\textbf{\tablenum[table-format=1.4]{0.7426}}} \\
 &  & 5 & \tablenum[table-format=1.4]{0.7847} & \tablenum[table-format=1.4]{0.0037} & \tablenum[table-format=1.4]{0.7856} &  \\
 &  & 10 & \tablenum[table-format=1.4]{0.7820} & \tablenum[table-format=1.4]{0.0074} & \tablenum[table-format=1.4]{0.7833} &  \\
 &  & 100 & \tablenum[table-format=1.4]{0.7828} & \tablenum[table-format=1.4]{0.0738} & \tablenum[table-format=1.4]{0.7831} &  \\
\cline{2-7}
 & \multirow{4}{*}{\makecell[l]{\textbf{CAMEL-CLIP}\\\textbf{(Ours)}}} & 1 & \textbf{\tablenum[table-format=1.4]{0.8182}} & \textbf{\tablenum[table-format=1.4]{0.0008}} & \textbf{\tablenum[table-format=1.4]{0.8182}} & \multirow{4}{*}{\tablenum[table-format=1.4]{0.7316}} \\
 &  & 5 & \textbf{\tablenum[table-format=1.4]{0.8136}} & \textbf{\tablenum[table-format=1.4]{0.0039}} & \textbf{\tablenum[table-format=1.4]{0.8143}} &  \\
 &  & 10 & \textbf{\tablenum[table-format=1.4]{0.8132}} & \textbf{\tablenum[table-format=1.4]{0.0077}} & \textbf{\tablenum[table-format=1.4]{0.8138}} &  \\
 &  & 100 & \textbf{\tablenum[table-format=1.4]{0.7997}} & \textbf{\tablenum[table-format=1.4]{0.0760}} & \textbf{\tablenum[table-format=1.4]{0.8027}} &  \\
\hline
\multirow{8}{*}{Gender} & \multirow{4}{*}{EEG-CLIP \cite{ndir2025eeg}} & 1 & \tablenum[table-format=1.4]{0.5307} & \tablenum[table-format=1.4]{0.0005} & \tablenum[table-format=1.4]{0.5307} & \multirow{4}{*}{\tablenum[table-format=1.4]{0.5166}} \\
 &  & 5 & \tablenum[table-format=1.4]{0.5264} & \tablenum[table-format=1.4]{0.0026} & \tablenum[table-format=1.4]{0.5271} &  \\
 &  & 10 & \tablenum[table-format=1.4]{0.5253} & \tablenum[table-format=1.4]{0.0051} & \tablenum[table-format=1.4]{0.5261} &  \\
 &  & 100 & \tablenum[table-format=1.4]{0.5180} & \tablenum[table-format=1.4]{0.0503} & \tablenum[table-format=1.4]{0.5195} &  \\
\cline{2-7}
 & \multirow{4}{*}{\makecell[l]{\textbf{CAMEL-CLIP}\\\textbf{(Ours)}}} & 1 & \textbf{\tablenum[table-format=1.4]{0.6316}} & \textbf{\tablenum[table-format=1.4]{0.0006}} & \textbf{\tablenum[table-format=1.4]{0.6316}} & \multirow{4}{*}{\textbf{\tablenum[table-format=1.4]{0.5787}}} \\
 &  & 5 & \textbf{\tablenum[table-format=1.4]{0.6394}} & \textbf{\tablenum[table-format=1.4]{0.0031}} & \textbf{\tablenum[table-format=1.4]{0.6381}} &  \\
 &  & 10 & \textbf{\tablenum[table-format=1.4]{0.6365}} & \textbf{\tablenum[table-format=1.4]{0.0061}} & \textbf{\tablenum[table-format=1.4]{0.6365}} &  \\
 &  & 100 & \textbf{\tablenum[table-format=1.4]{0.6180}} & \textbf{\tablenum[table-format=1.4]{0.0594}} & \textbf{\tablenum[table-format=1.4]{0.6214}} &  \\
\hline
\multirow{8}{*}{Age} & \multirow{4}{*}{EEG-CLIP \cite{ndir2025eeg}} & 1 & \tablenum[table-format=1.4]{0.5999} & \tablenum[table-format=1.4]{0.0006} & \tablenum[table-format=1.4]{0.5999} & \multirow{4}{*}{\tablenum[table-format=1.4]{0.5544}} \\
 &  & 5 & \tablenum[table-format=1.4]{0.5882} & \tablenum[table-format=1.4]{0.0028} & \tablenum[table-format=1.4]{0.5904} &  \\
 &  & 10 & \tablenum[table-format=1.4]{0.5760} & \tablenum[table-format=1.4]{0.0056} & \tablenum[table-format=1.4]{0.5811} &  \\
 &  & 100 & \tablenum[table-format=1.4]{0.5614} & \tablenum[table-format=1.4]{0.0543} & \tablenum[table-format=1.4]{0.5647} &  \\
\cline{2-7}
 & \multirow{4}{*}{\makecell[l]{\textbf{CAMEL-CLIP}\\\textbf{(Ours)}}} & 1 & \textbf{\tablenum[table-format=1.4]{0.6938}} & \textbf{\tablenum[table-format=1.4]{0.0007}} & \textbf{\tablenum[table-format=1.4]{0.6938}} & \multirow{4}{*}{\textbf{\tablenum[table-format=1.4]{0.5965}}} \\
 &  & 5 & \textbf{\tablenum[table-format=1.4]{0.6712}} & \textbf{\tablenum[table-format=1.4]{0.0032}} & \textbf{\tablenum[table-format=1.4]{0.6758}} &  \\
 &  & 10 & \textbf{\tablenum[table-format=1.4]{0.6632}} & \textbf{\tablenum[table-format=1.4]{0.0064}} & \textbf{\tablenum[table-format=1.4]{0.6687}} &  \\
 &  & 100 & \textbf{\tablenum[table-format=1.4]{0.6390}} & \textbf{\tablenum[table-format=1.4]{0.0616}} & \textbf{\tablenum[table-format=1.4]{0.6441}} &  \\
\hline
\end{tabular}}
\end{table}

\clearpage
\begin{figure}[H]
    \centering
    \includegraphics[width=1.0\linewidth]{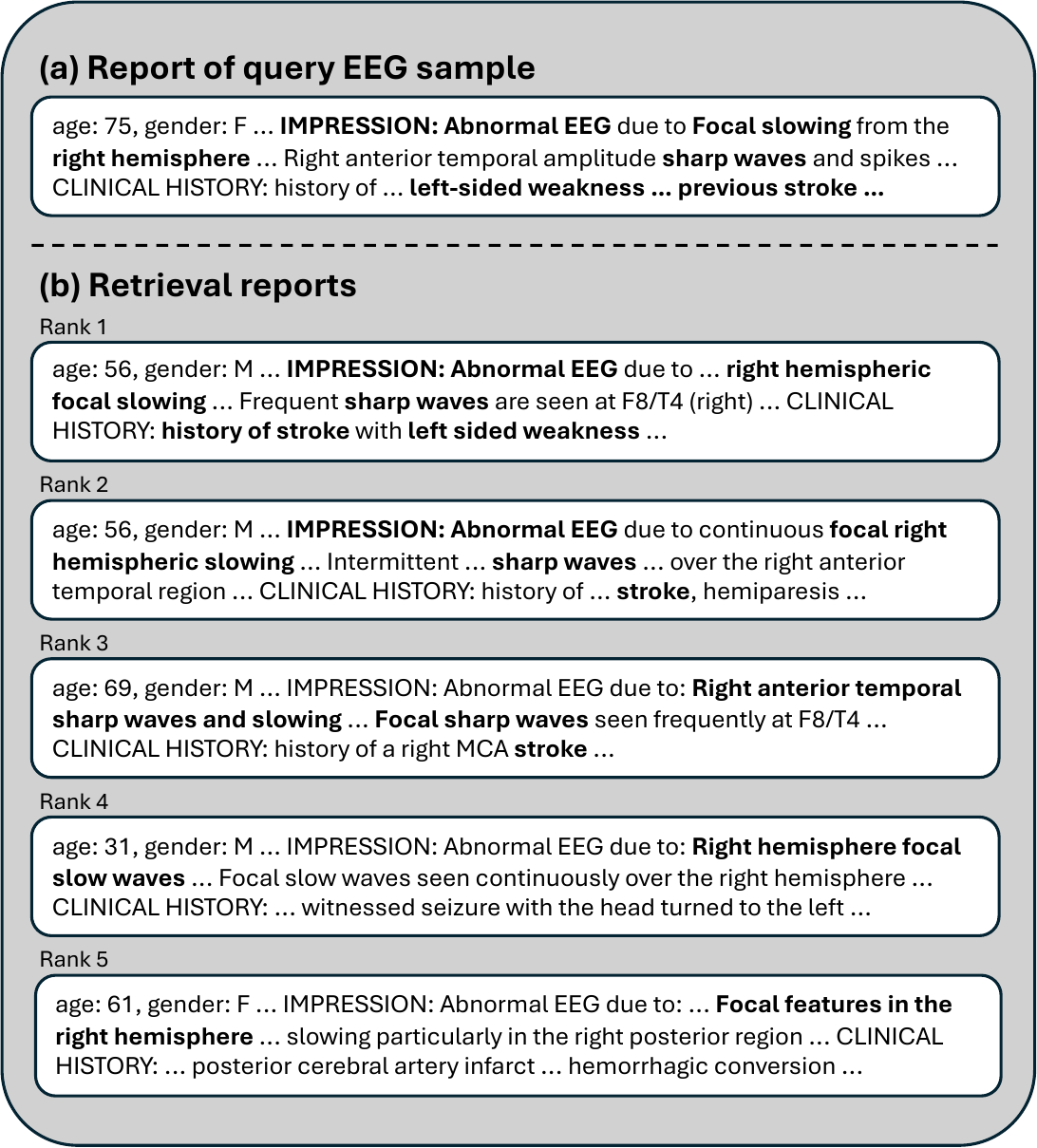}
    \caption{\textbf{Sample of EEG-text retrieval from the TUAB dataset.} (a) Text report of query EEG sample. (b) Retrieval report. }
    \label{fig10}
\end{figure}

Figure \ref{fig10} presents the actual results of the EEG-text retrieval task in the TUAB dataset conducted by CAMEL-CLIP. Figure \ref{fig10}(a) illustrates the original report of the EEG sample used as a query, while Figure \ref{fig10}(b) displays the training set reports retrieved by the model and ranked by similarity. A comparison between Figure \ref{fig10}(a) and (b) reveals that key clinical terms such as ``Right hemisphere focal slowing'', ``sharp waves'', and ``stroke'' appear consistently in both.

Overall, the comparison with EEG-only pretraining suggests the contribution of the text modality to downstream performance. The higher cosine similarity of positive EEG--text pairs and the improved retrieval performance further support alignment between EEG and text representations. These results suggest that the proposed methods help improve EEG--text alignment and enable the model to benefit from text information during pretraining.

% Response Table A5

\FloatBarrier
\subsection{Computational overhead analysis}
\label{sec:overhead}

In Experiment 6, we evaluate the computational overhead associated with channel-wise projection and variable-length embeddings by comparing conditions with and without DLA, which combines DCP and DCL. Because DCP and DCL are designed to be used together, we evaluate the computational overhead of their combined configuration. We measure training time, as well as inference latency, throughput, peak allocated GPU memory, parameter count, and supported-operator FLOP estimates for EEG embedding generation. We use the TUAB dataset for training.

As shown in Table \ref{tab:overhead}, CAMEL-CLIP shows only marginal increases in parameter count, GFLOPs per sample, latency, and GPU memory usage, while maintaining throughput comparable to that of the baseline. The larger embedding dimensionality reflects its design to preserve channel-specific features, allowing individual channel representations to be examined separately and supporting model interpretability.

During training, however, DLA introduces an additional computational cost, primarily due to DCL. In channel-level contrastive learning, $N$ channel representations are sampled from each of the $B$ EEG samples in a mini-batch, increasing the number of embeddings used for contrastive learning from $B$ to $BN$. For example, with $B = 512$ and $N = 3$, 1,536 embeddings are used to compute the CCL loss. As a result, enabling DLA in CAMEL-CLIP increases the total training time from approximately 3.06 hours to approximately 9.07 hours in our measurement environment. Although training time increases, the ablation study demonstrates substantial performance improvements when DLA is used.

% Response Table A14
\clearpage
\begin{table}[H]
\centering
\small
\setlength{\tabcolsep}{3pt}
\renewcommand{\arraystretch}{1.35}
\caption{\textbf{Computational overhead of DLA.} We compare models trained without and with DLA. Total training time is reported in hours and measured during pretraining with the corresponding loss configuration, including DCL in CAMEL-CLIP. GFLOPs per sample, latency, throughput, and GPU memory usage are measured during EEG embedding inference. Parameter count and embedding dimension refer to the EEG branch. Latency is measured with batch size 1 and reported as mean $\pm$ SD; throughput and GPU memory are measured with batch size 64. DLA denotes the combined configuration of dynamic channel projection (DCP) and dual-level contrastive learning (DCL).}
\label{tab:overhead}
\begin{tabular*}{\linewidth}{@{\extracolsep{\fill}}rrr@{}}
\hline
\textbf{Setting} & \multicolumn{1}{r}{\makecell[r]{\textbf{without DLA}}} & \multicolumn{1}{r}{\makecell[r]{\textbf{with DLA}}} \\
\hline
\#Params & 5,302,084 & 5,355,148 \\

GFLOPs / sample & 0.635550 & 0.636659 \\

Latency (ms) & 10.958 $\pm$ 0.049 & 11.046 $\pm$ 0.048 \\

Throughput (samples/s) & 5,778 & 5,702 \\

GPU memory (MiB) & 84.78 & 84.98 \\

Embedding dim. & 64 & 1,408 \\

Total training time (h) & $\approx$ 3.06 & $\approx$ 9.07 \\
\hline
\end{tabular*}
\end{table}

\FloatBarrier
\section{Conclusion}
\label{sec6}

We introduce CAMEL-CLIP, a multimodal EEG–text foundation model designed to maintain robustness under varying channel configurations. To explicitly address channel heterogeneity, CAMEL-CLIP incorporates three key components: (1) CAPE, which represents channel identity using anatomical attributes and reference type; (2) DCP, which constructs channel-specific text representations across varying channel configurations; and (3) DCL, which aligns EEG and text representations at both the channel and sample levels.

By preserving channel identity and incorporating channel-specific information into EEG–text alignment, CAMEL-CLIP learns representations that transfer effectively across datasets. It achieves the highest mean balanced accuracy among the evaluated baselines under linear probing on the CHB-MIT and MUMTAZ-2016 datasets. The complementary relationship between CAPE and Dual-level Alignment supports this downstream performance. Through CCL, the EEG encoder learns consistent and distinct channel representations, including for channels excluded from pretraining. CAMEL-CLIP also maintains high classification performance when the channel configuration changes without classifier retraining, supporting its ability to handle channel heterogeneity.

Beyond downstream classification, CAMEL-CLIP benefits from text information during pretraining and learns correspondence between EEG signals and textual descriptions. The higher similarity of corresponding EEG–text pairs and improved retrieval of reports matching the evaluated attributes support the effectiveness of the proposed methods for EEG–text alignment. These capabilities offer the potential to provide contextual information for EEG interpretation.

Despite these promising results, multimodal EEG–text modeling continues to face a fundamental challenge: the limited availability of large-scale, high-quality paired datasets that fully capture the complexity of EEG signals. Consequently, future work should more systematically investigate strategies to mitigate representation degradation under limited paired data. Potential directions include leveraging additional modalities to provide richer supervisory signals, developing EEG-specific augmentation schemes, and designing more principled pairing and alignment algorithms between EEG and text. Progress along these lines may further enhance the robustness of multimodal EEG-text models.

\section*{Code and Data Availability}
The source code and pretrained model weights for CAMEL-CLIP are publicly available at \url{https://github.com/FrogLlama/CAMEL-CLIP} to facilitate reproduction of the proposed method. The datasets used in this study are described and referenced in Section~\ref{sec4}.

\section*{CRediT authorship contribution statement}
\textbf{Hanseul Choi}: Conceptualization, Methodology, Software, Visualization, Writing - original draft. \textbf{Jinyeong Park}: Conceptualization, Methodology, Writing - review \& editing. \textbf{Seongwon Jin}: Conceptualization, Visualization, Writing - review \& editing. \textbf{Sungho Park}: Supervision, Writing - review \& editing. \textbf{Jibum Kim}: Conceptualization, Supervision, Project administration, Funding acquisition, Writing - review \& editing.

\section*{Funding}
This work was supported by research funding from the Incheon National University Artificial Intelligence \& Big Data Center in 2025.

\section*{Declaration of competing interest}
The authors declare that they have no known competing financial interests or personal relationships that could have appeared to influence the work reported in this paper.

\section*{Declaration of generative AI and AI-assisted technologies in the manuscript preparation process}
During the preparation of this manuscript, the authors used generative AI tools to assist with language editing and clarity improvement. The authors reviewed and revised the content and take full responsibility for the accuracy, integrity, and originality of the work.

\bibliographystyle{elsarticle-num}
\bibliography{refs}

\begin{thebibliography}{10}
\expandafter\ifx\csname url\endcsname\relax
  \def\url#1{\texttt{#1}}\fi
\expandafter\ifx\csname urlprefix\endcsname\relax\def\urlprefix{URL }\fi
\expandafter\ifx\csname href\endcsname\relax
  \def\href#1#2{#2} \def\path#1{#1}\fi

\bibitem{ay2019automated}
B.~Ay, O.~Yildirim, M.~Talo, U.~B. Baloglu, G.~Aydin, S.~D. Puthankattil, U.~R.
  Acharya, Automated depression detection using deep representation and
  sequence learning with {EEG} signals, Journal of Medical Systems 43~(7)
  (2019) 205.
\newblock \href {https://doi.org/10.1007/s10916-019-1345-y}
  {\path{doi:10.1007/s10916-019-1345-y}}.

\bibitem{shoeibi2021epileptic}
A.~Shoeibi, M.~Khodatars, N.~Ghassemi, M.~Jafari, P.~Moridian, R.~Alizadehsani,
  F.~Khozeimeh, A.~Zare, H.~Hosseini-Nejad, K.~Bahaadinbeigy, D.~Sadeghi,
  T.~Naghibi, M.~Jafari, S.~Hussain, U.~R. Acharya, Epileptic seizures
  detection using deep learning techniques: a review, International Journal of
  Environmental Research and Public Health 18 (2021) 5780.
\newblock \href {https://doi.org/10.3390/ijerph18115780}
  {\path{doi:10.3390/ijerph18115780}}.

\bibitem{solis2024machine}
T.~Solis-Escalante, A.~Tapia-Gallardo, E.~Y. de~la Hoz-Franco, P.~Gomez-Gil,
  J.~P. Amezquita-Sanchez, Machine and deep learning trends in {EEG}-based
  detection and diagnosis of {Alzheimer}'s disease: a systematic review,
  Engineering Proceedings 68~(1) (2024) 78.

\bibitem{lawhern2018eegnet}
V.~J. Lawhern, A.~J. Solon, N.~R. Waytowich, S.~M. Gordon, C.~P. Hung, B.~J.
  Lance, {EEGNet}: a compact convolutional neural network for {EEG}-based
  brain--computer interfaces, Journal of Neural Engineering 15~(5) (2018)
  056013.
\newblock \href {https://doi.org/10.1088/1741-2552/aace8c}
  {\path{doi:10.1088/1741-2552/aace8c}}.

\bibitem{song2022eeg}
Y.~Song, Q.~Zheng, B.~Liu, X.~Gao, {EEG} conformer: convolutional transformer
  for {EEG} decoding and visualization, IEEE Transactions on Neural Systems and
  Rehabilitation Engineering 31 (2023) 710--719.
\newblock \href {https://doi.org/10.1109/TNSRE.2022.3230250}
  {\path{doi:10.1109/TNSRE.2022.3230250}}.

\bibitem{li2022motor}
H.~Li, M.~Ding, R.~Zhang, C.~Xiu, Motor imagery {EEG} classification algorithm
  based on {CNN-LSTM} feature fusion network, Biomedical Signal Processing and
  Control 72 (2022) 103342.
\newblock \href {https://doi.org/10.1016/j.bspc.2021.103342}
  {\path{doi:10.1016/j.bspc.2021.103342}}.

\bibitem{saha2020intra}
S.~Saha, M.~Baumert, Intra- and inter-subject variability in {EEG}-based
  sensorimotor brain computer interface: a review, Frontiers in Computational
  Neuroscience 13 (2020) 87.
\newblock \href {https://doi.org/10.3389/fncom.2019.00087}
  {\path{doi:10.3389/fncom.2019.00087}}.

\bibitem{jiang2024large}
W.~Jiang, L.~Zhao, B.-L. Lu, Large brain model for learning generic
  representations with tremendous {EEG} data in {BCI}, in: The Twelfth
  International Conference on Learning Representations, 2024.

\bibitem{wang2024cbramod}
J.~Wang, S.~Zhao, Z.~Luo, Y.~Zhou, H.~Jiang, S.~Li, T.~Li, G.~Pan, {CBraMod}: A
  criss-cross brain foundation model for {EEG} decoding, arXiv preprint
  arXiv:2412.07236 (2024).

\bibitem{xiong2025eeg}
W.~Xiong, J.~Li, J.~Li, K.~Zhu, {EEG-FM-Bench}: a comprehensive benchmark for
  the systematic evaluation of {EEG} foundation models, arXiv preprint
  arXiv:2508.17742 (2025).

\bibitem{zhang2025brain}
X.~Zhou, C.~Liu, Z.~Chen, K.~Wang, Y.~Ding, Z.~Jia, Q.~Wen, Brain foundation
  models: a survey on advancements in neural signal processing and brain
  discovery, arXiv preprint arXiv:2503.00580 (2025).

\bibitem{kuruppu2025eeg}
G.~Kuruppu, N.~Wagh, V.~Kremen, S.~Pati, G.~Worrell, Y.~Varatharajah, {EEG}
  foundation models: a critical review of current progress and future
  directions, arXiv preprint arXiv:2507.11783 (2025).

\bibitem{chen2025hear}
Z.~Chen, C.~Qin, W.~You, R.~Liu, C.~Chu, R.~Yang, K.~C. Tan, J.~Wu, {HEAR}: An
  {EEG} foundation model with heterogeneous electrode adaptive representation,
  arXiv preprint arXiv:2510.12515 (2025).

\bibitem{yao2019which}
D.~Yao, Y.~Qin, S.~Hu, L.~Dong, M.~L.~B. Vega, P.~A.~V. Sosa, Which reference
  should we use for {EEG} and {ERP} practice?, Brain Topography 32~(4) (2019)
  530--549.
\newblock \href {https://doi.org/10.1007/s10548-019-00707-x}
  {\path{doi:10.1007/s10548-019-00707-x}}.

\bibitem{ndir2025eeg}
T.~C. Ndir, R.~T. Schirrmeister, T.~Ball, {EEG-CLIP}: learning {EEG}
  representations from natural language descriptions, Frontiers in Robotics and
  AI 12 (2025) 1625731.
\newblock \href {https://doi.org/10.3389/frobt.2025.1625731}
  {\path{doi:10.3389/frobt.2025.1625731}}.

\bibitem{yang2023biot}
C.~Yang, M.~B. Westover, J.~Sun, {BIOT}: Biosignal transformer for cross-data
  learning in the wild, in: Advances in Neural Information Processing Systems,
  Vol.~36, 2023, pp. 78240--78260.

\bibitem{yan2025cross}
R.~Yan, Y.~Li, H.~Ding, F.~Wang, Cross-domain {EEG}-based emotion recognition
  with contrastive learning, arXiv preprint arXiv:2511.05293 (2025).

\bibitem{radford2021learning}
A.~Radford, J.~W. Kim, C.~Hallacy, A.~Ramesh, G.~Goh, S.~Agarwal, G.~Sastry,
  A.~Askell, P.~Mishkin, J.~Clark, G.~Krueger, I.~Sutskever, Learning
  transferable visual models from natural language supervision, in: Proceedings
  of the 38th International Conference on Machine Learning, Vol. 139 of
  Proceedings of Machine Learning Research, PMLR, 2021, pp. 8748--8763.

\bibitem{patil2024coordconformer}
S.~Patil, R.~T. Schirrmeister, F.~Hutter, T.~Ball, {CoordConformer}:
  Heterogenous {EEG} datasets decoding using transformers, in: {ICML} 2024
  Workshop on Geometry-grounded Representation Learning and Generative
  Modeling, 2024, extended abstract.

\bibitem{yi2023learning}
K.~Yi, Y.~Wang, K.~Ren, D.~Li, Learning topology-agnostic {EEG} representations
  with geometry-aware modeling, in: Advances in Neural Information Processing
  Systems, Vol.~36, 2023, pp. 53875--53891.

\bibitem{doner2025luna}
B.~Döner, T.~M. Ingolfsson, L.~Benini, Y.~Li, {LUNA}: Efficient and
  topology-agnostic foundation model for {EEG} signal analysis, arXiv preprint
  arXiv:2510.22257 (2025).
\newblock \href {https://doi.org/10.48550/arXiv.2510.22257}
  {\path{doi:10.48550/arXiv.2510.22257}}.

\bibitem{huang2019clinicalbert}
K.~Huang, J.~Altosaar, R.~Ranganath, {ClinicalBERT}: modeling clinical notes
  and predicting hospital readmission, arXiv preprint arXiv:1904.05342 (2019).

\bibitem{devlin2019bert}
J.~Devlin, M.-W. Chang, K.~Lee, K.~Toutanova, {BERT}: Pre-training of deep
  bidirectional transformers for language understanding, in: Proceedings of the
  2019 Conference of the North American Chapter of the Association for
  Computational Linguistics: Human Language Technologies, Volume 1 (Long and
  Short Papers), Association for Computational Linguistics, Minneapolis,
  Minnesota, 2019, pp. 4171--4186.
\newblock \href {https://doi.org/10.18653/v1/N19-1423}
  {\path{doi:10.18653/v1/N19-1423}}.

\bibitem{jasper1958ten}
H.~H. Jasper, The ten-twenty electrode system of the international federation,
  Electroencephalography and Clinical Neurophysiology 10 (1958) 371--375.

\bibitem{oord2018representation}
A.~van~den Oord, Y.~Li, O.~Vinyals, Representation learning with contrastive
  predictive coding, arXiv preprint arXiv:1807.03748 (2018).

\bibitem{obeid2016temple}
I.~Obeid, J.~Picone, The temple university hospital {EEG} data corpus,
  Frontiers in Neuroscience 10 (2016) 196.
\newblock \href {https://doi.org/10.3389/fnins.2016.00196}
  {\path{doi:10.3389/fnins.2016.00196}}.

\bibitem{shoeb2009application}
A.~H. Shoeb, Application of machine learning to epileptic seizure onset
  detection and treatment, Ph.D. thesis, Massachusetts Institute of Technology
  (2009).

\bibitem{goldberger2000physiobank}
A.~L. Goldberger, L.~A.~N. Amaral, L.~Glass, J.~M. Hausdorff, P.~C. Ivanov,
  R.~G. Mark, J.~E. Mietus, G.~B. Moody, C.-K. Peng, H.~E. Stanley,
  {PhysioBank}, {PhysioToolkit}, and {PhysioNet}: components of a new research
  resource for complex physiologic signals, Circulation 101~(23) (2000)
  e215--e220.
\newblock \href {https://doi.org/10.1161/01.CIR.101.23.e215}
  {\path{doi:10.1161/01.CIR.101.23.e215}}.

\bibitem{guttag2010chb}
J.~Guttag, {CHB-MIT Scalp EEG Database}, PhysioNetVersion 1.0.0 (2010).
\newblock \href {https://doi.org/10.13026/C2K01R} {\path{doi:10.13026/C2K01R}}.

\bibitem{mumtaz2016mdd}
W.~Mumtaz, {MDD} patients and healthy controls {EEG} data (new), figshare,
  dataset, Version 2 (2016).
\newblock \href {https://doi.org/10.6084/m9.figshare.4244171.v2}
  {\path{doi:10.6084/m9.figshare.4244171.v2}}.

\bibitem{ma2026codebrain}
J.~Ma, et~al., {CodeBrain}: Bridging decoupled tokenizer and multi-scale
  architecture for {EEG} foundation model, in: International Conference on
  Learning Representations, 2026.

\bibitem{sirca2026beyond}
U.~Širca, et~al., Beyond accuracy: Robustness, interpretability and
  expressiveness of {EEG} foundation models, arXiv preprint arXiv:2605.17562
  (2026).

\bibitem{mcinnes2018umap}
L.~McInnes, J.~Healy, J.~Melville, {UMAP}: Uniform manifold approximation and
  projection for dimension reduction, arXiv preprint arXiv:1802.03426 (2018).

\bibitem{friedman2023vendi}
D.~Friedman, A.~B. Dieng, The vendi score: A diversity evaluation metric for
  machine learning, Transactions on Machine Learning Research (2023).

\bibitem{rousseeuw1987silhouettes}
P.~J. Rousseeuw, Silhouettes: A graphical aid to the interpretation and
  validation of cluster analysis, Journal of Computational and Applied
  Mathematics 20 (1987) 53--65.

\bibitem{thakur2021beir}
N.~Thakur, N.~Reimers, A.~Rücklé, A.~Srivastava, I.~Gurevych, {BEIR}: A
  heterogeneous benchmark for zero-shot evaluation of information retrieval
  models, in: Proceedings of the Neural Information Processing Systems Track on
  Datasets and Benchmarks, Vol.~1, 2021.

\bibitem{khattab2020colbert}
O.~Khattab, M.~Zaharia, {ColBERT}: Efficient and effective passage search via
  contextualized late interaction over {BERT}, in: Proceedings of the 43rd
  International ACM SIGIR Conference on Research and Development in Information
  Retrieval, Association for Computing Machinery, New York, NY, USA, 2020, pp.
  39--48.
\newblock \href {https://doi.org/10.1145/3397271.3401075}
  {\path{doi:10.1145/3397271.3401075}}.

\bibitem{santhanam2022colbertv2}
K.~Santhanam, O.~Khattab, J.~Saad-Falcon, C.~Potts, M.~Zaharia, {ColBERTv2}:
  Effective and efficient retrieval via lightweight late interaction, in:
  Proceedings of the 2022 Conference of the North American Chapter of the
  Association for Computational Linguistics: Human Language Technologies,
  Association for Computational Linguistics, Seattle, United States, 2022, pp.
  3715--3734.
\newblock \href {https://doi.org/10.18653/v1/2022.naacl-main.272}
  {\path{doi:10.18653/v1/2022.naacl-main.272}}.

\end{thebibliography}

\end{document}